\documentclass[10pt,twocolumn,letterpaper]{article}
\pdfoutput=1

\usepackage{cvpr}
\usepackage{times}
\usepackage{epsfig}
\usepackage{graphicx}
\usepackage{amsmath}
\usepackage{amssymb}

\usepackage{caption}
\usepackage{subcaption}
\usepackage{tablefootnote}
\usepackage{multirow}
\usepackage{hhline}

\graphicspath{{figs/}}

\newcommand{\Mark}[1]{\textsuperscript#1}

\usepackage[pagebackref=true,breaklinks=true,letterpaper=true,colorlinks,bookmarks=false]{hyperref}

\cvprfinalcopy 

\def\cvprPaperID{1637} 

\ifcvprfinal\fi
\begin{document}

\title{Dense Intrinsic Appearance Flow for Human Pose Transfer}

\author{
	\makebox[\linewidth][c]{Yining Li\Mark{1}\hspace{1.5em}Chen Huang\Mark{2}\hspace{1.5em}Chen Change Loy\Mark{3}}\\
	\Mark{1}CUHK-SenseTime Joint Lab, The Chinese University of Hong Kong\\
	\Mark{2}Robotics Institute, Carnegie Mellon University\\
	\Mark{3}School of Computer Science and Engineering, Nanyang Technological University\\	
	{\tt\small ly015@ie.cuhk.edu.hk~~chenh2@andrew.cmu.edu~~ccloy@ntu.edu.sg}
}


\maketitle


\begin{abstract}
We present a novel approach for the task of human pose transfer, which aims at synthesizing a new image of a person from an input image of that person and a target pose. 
%
%
Unlike existing methods, we propose to estimate dense and intrinsic 3D appearance flow to better guide the transfer of pixels between poses. In particular, we wish to generate the 3D flow from just the reference and target poses. Training a network for this purpose is non-trivial, especially when the annotations for 3D appearance flow are scarce by nature. 
We address this problem through a flow synthesis stage. This is achieved by fitting a 3D model to the given pose pair and project them back to the 2D plane to compute the dense appearance flow for training. The synthesized ground-truths are then used to train a feedforward network for efficient mapping from the input and target skeleton poses to the 3D appearance flow. 
With the appearance flow, we perform feature warping on the input image and generate a photorealistic image of the target pose. 
Extensive results on DeepFashion and Market-1501 datasets demonstrate the effectiveness of our approach over existing methods.
Our code is available at \url{http://mmlab.ie.cuhk.edu.hk/projects/pose-transfer/}

\end{abstract}
\begin{figure*}[t]
	\centering
	\includegraphics[width=\linewidth]{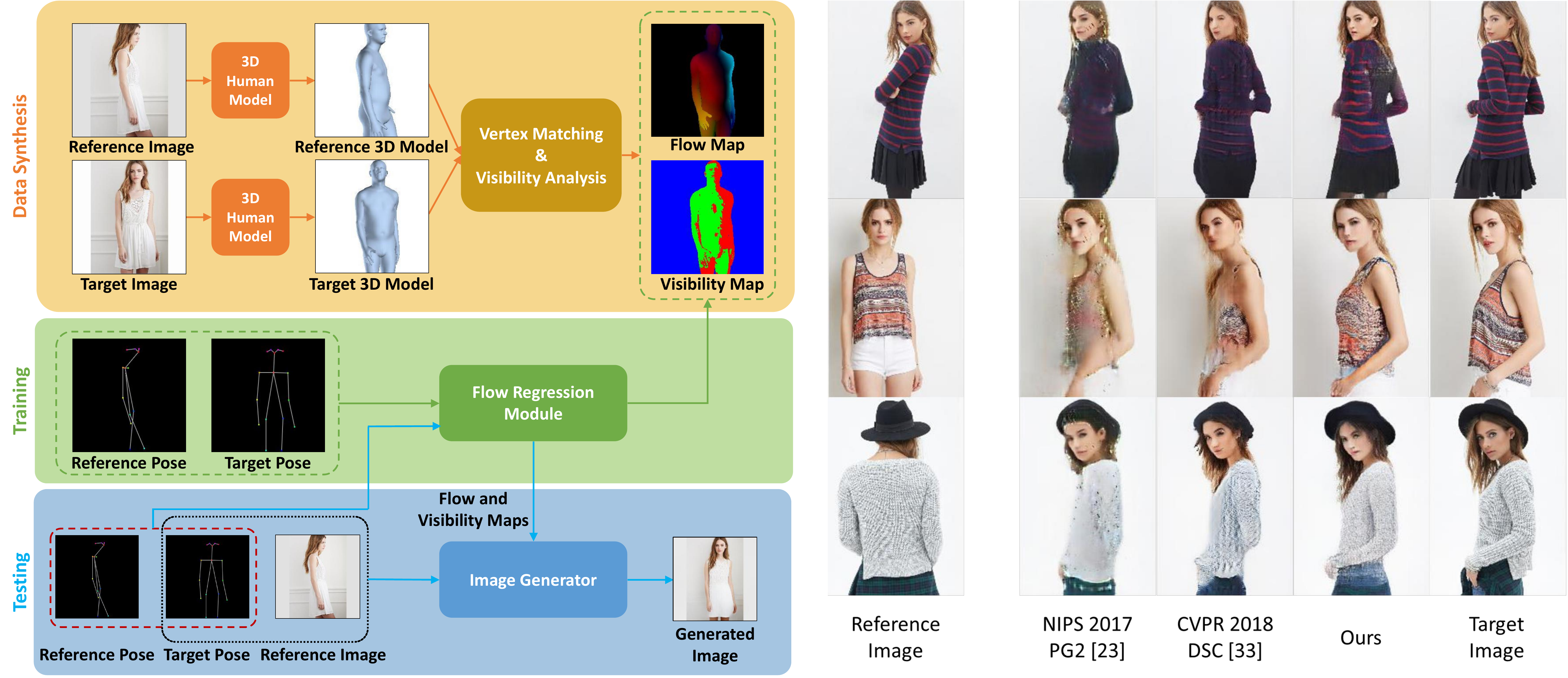}
	\caption{The proposed human pose transfer method with dense intrinsic 3D appearance flow generates higher quality images in comparison to baselines. (\textbf{Left}) The core of our method is a flow regression module (the green box) that can transform the reference and target poses into a 3D appearance flow map and a visibility map.}
	\label{fig:intro}
\end{figure*}

\section{Introduction}
\label{sec:introduction}

The ability to predict what an object will look like from a new viewpoint is fundamental to intelligence. Human pose transfer~\cite{neverova2018dense} is an important instantiation of such view synthesis task. Given a single view/pose of one person, the goal is to synthesize an image of that person in arbitrary poses. This task is of great value to a wide range of applications in computer vision and graphics. Examples include video synthesis and editing and data augmentation for problems like person re-identification where it is hard to acquire enough same-person images from different cameras.

Despite the rapid progress in deep generative models like Generative Adversarial Networks (GAN)~\cite{goodfellow2014generative} and Variational Auto Encoders (VAE)~\cite{kingma2013auto}, human image generation between poses is still exceedingly difficult. The main challenge is to model the large variations in 2D appearance due to the change in 3D pose. This is further compounded by human body self-occlusion that induces ambiguities in inferring unobserved pixels for the target pose. In general, successful human pose transfer requires a good representation or disentangling of human pose and appearance, which is non-trivial to learn from data. The ability to infer invisible parts is also necessary. Moreover, the image visual quality largely depends on whether the high frequency details can be preserved,~\eg~in cloth or face regions.



Most existing methods for human pose transfer~\cite{balakrishnan2018synthesizing,esser2018variational,lassner2017generative,ma2017pose,ma2018disentangled,pumarola2018unsupervised,raj2018swapnet,zhao2018multi} employ an encoder-decoder architecture to learn the appearance transformation from an input image, guided by the input and target 2D pose encoded with some keypoints of the human-body joints. However, such keypoint-based representation is only able to capture rough spatial deformations, but not fine-grained ones. As a result, distortions or unrealistic details are often produced, especially in the presence of large pose change with non-rigid body deformations. Recent advances either decompose the overall deformation by a set of local affine transformations~\cite{siarohin2018deformable}, or use a more detailed pose representation than the keypoint-based one. The latter is to enable `dense appearance flow' computation that more accurately specifies how to move pixels from the input pose. Neverova~\etal~\cite{neverova2018dense} showed that the surface-based pose representation via DensePose~\cite{guler2018densepose} serves as a better alternative. Zanfir~\etal~\cite{zanfir2018human} turned to fit a 3D model to both input and target images, and then perform appearance transfer between the corresponding vertices. The resulting appearance flow with 3D geometry supervision is more ideal, but the 3D model fitting would incur too much burden at inference time. 



In this paper, we propose a novel approach to human pose transfer that integrates implicit reasoning about 3D geometry from 2D representations only. This allows us to share the benefits of using 3D geometry for accurate pose transfer but at much faster speed. Our key idea is to recover from training image pairs (along with their pose keypoints) the underlying 3D models, which when projected back to 2D image plane can provide the ground-truth appearance flow for us to learn from. Such \textit{dense and intrinsic appearance flow} implicitly encodes the 3D structures of human body. Then we train an appearance flow generation module, represented by the traditional feedforward network, which directly regresses from a pair of 2D poses to the corresponding appearance flow. This module helps us to bypass the expensive 3D model fitting at test time, and predict the intrinsic pixel-wise correspondence pretty fast without requiring explicit access to 3D geometry.

Figure~\ref{fig:intro} (left) illustrates our overall image generation framework. Given a reference image (and its pose) and the target pose, we first use a variant of U-Net~\cite{ronneberger2015u} to encode the image and target pose respectively. Then our appearance flow module generates a 3D flow map from the pose pair, and further generates a visibility map to account for the missing pixels in the target pose due to self-occlusions. The visibility map proves necessary for our network to synthesize missing pixels at the correct locations. To render the final image in target pose, the encoded image features are first warped through the generated flow map, and then passed to a gating module guided by the visibility map. Finally, our pose decoder concatenates such processed image features to generate the image. Our U-Net-type image generator and appearance flow module are trained end-to-end so as to optimize a combination of reconstruction, adversarial and perceptual losses. Our approach is able to generate high quality images on DeepFashion~\cite{liu2016deepfashion} and Market-1501~\cite{zheng2015scalable} datasets, showing consistent improvements over existing image generators based on keypoints or other pose representations. Our method also achieves compelling quantitative results.

The main contributions of this paper can be summarized as follows:
\begin{itemize}
	\vspace{-0.5em}
	\item[$\bullet$] A feedforward appearance flow generation module is proposed to efficiently encode the dense and intrinsic correspondences in 3D space for human pose transfer.
	\vspace{-1.5em}
	\item[$\bullet$] An end-to-end image generation framework is learned to move pixels with the appearance flow map and handle self-occlusions with a visibility map.
	\vspace{-0.5em} 
	\item[$\bullet$] State-of-the-art performance and high quality images are produced on DeepFashion dataset.
\end{itemize}
\section{Related Work}
\label{sec:related_work}
\noindent
\textbf{Deep generative image models.} Recent years have seen a breakthrough of deep generative methods for image generation, using Generative Adversarial Networks (GAN)~\cite{goodfellow2014generative}, Variational Autoencoder (VAE)~\cite{kingma2013auto} and so on. Among these, GAN has drawn a great attention due to its capability of generating realistic images. Follow-up works make GANs conditional, generating images based on extra inputs like class labels~\cite{mirza2014conditional}, natural language descriptions~\cite{zhu2017be,zhang2017stackgan++,zhang2017stackgan} or images from another domain~\cite{pix2pix2017} that leads to an image-to-image domain transfer framework. Adversarial learning has also shown its effectiveness in many other tasks like image super-resolution~\cite{ledig2017photo,wang2018recovering,wang2018esrgan} and texture generation~\cite{xian2016texturegan}.


\noindent
\textbf{Human pose transfer.} Generating human-centric images is an important sub-area of image synthesis. Example tasks range from generating full human body in clothing~\cite{lassner2017generative} to generating human action sequences~\cite{chan2018everybody}. Ma~\etal~\cite{ma2017pose} are the first ones to approach the task of human pose transfer, which aims to generate a person image in a target pose if a reference image of that person is given beforehand. The pose comprised of 18 keypoints, is represented as a 18-channel keypoint heatmap. Then it is concatenated with the reference image and fed into a two-stage CNN for adversarial training. Zhao~\etal~\cite{zhao2018multi} adopted a similar coarse-to-fine approach to generate new images, but conditioned on the target view rather than target pose with multiple keypoints. To better handle the non-rigid body deformation in large pose transfer, Siarohin~\etal~\cite{siarohin2018deformable} proposed Deformable GAN to decompose the overall deformation by a set of local affine transformations. Another line of works~\cite{esser2018variational,ma2018disentangled,pumarola2018unsupervised,raj2018swapnet} focus on disentangling human appearance and pose with weak supervision. With only single image rather than a pair as input, these methods try to distill appearance information in a separate embedding, sometimes with the help of cycle-consistent penalty~\cite{pumarola2018unsupervised}.

\noindent
\textbf{Geometry-based pose transfer.} Some recent works integrate geometric constraints of human body to improve pose transfer. Neverova~\etal~\cite{neverova2018dense} proposed a surface-based pose representation on top of DensePose~\cite{guler2018densepose}. This allows to map the body pixels to a meaningful UV-coordinate space, where surface interpolation and inpainting can happen before warping back to the image space. Zanfir~\cite{zanfir2018human} on the other hand, proposed to leverage 3D human model to explicitly capture the body deformations. Specifically, they fit a 3D human model~\cite{loper2015smpl} to both source and target images using the method in~\cite{zanfir2018monocular}, where a human body is represented by 6890 surface vertices. Then the pixels on overlapping vertices are directly transfered to the target image, while the invisible vertices in source image are hallucinated using a neural network. The main drawback of this work is that 3D model fitting is computationally expensive and is not always accurate. Our method avoids the costly 3D model fitting at test time, and instead learns to predict the 2D appearance flow map and visibility map defined by 3D correspondences in order to guide pixel transfer. This enables implicit reasoning about 3D geometry without requiring access to it.

\noindent
\textbf{Appearance flow for view synthesis.} Optical flow~\cite{horn1981determining} provides dense pixel-to-pixel correspondence between two images, and has been proved useful in tasks like action recognition in video~\cite{simonyan2014two}. Appearance flow~\cite{zhou2016view} also specifies dense correspondence often between images with different view-points, which is closer to our setting. However, previous works mainly estimate appearance flow from simple view transformations (\eg,~a global rotation) or rigid objects (\eg,~a car). Whereas our appearance flow module deals with the articulated human body with arbitrary pose transformation.


\section{Methodology}
\label{sec:method}
\subsection{Problem Formulation and Notations}
\label{sec:notation}
Given a reference person image $x$ and a target pose $p$, our goal is to generate a photorealistic image $\hat{x}$ for that person but in pose $p$. For arbitrary pose transfer, we simply adopt the commonly-used pose representation to guide such transfer. Specifically, we use 18 human keypoints extracted by a pose estimator~\cite{cao2017realtime} as in~\cite{ma2017pose, siarohin2018deformable}. The keypoints are encoded into a 18-channel binary heatmap, where each channel is filled with 1 within a radius of 8 pixels around the corresponding keypoint and 0 elsewhere. During training, we consider the image pair $(x_1, x_2)$ (source and target) with their corresponding poses $(p_1, p_2)$. The model takes the triplet $(x_1, p_1, p_2)$ as inputs and tries to generate $\hat{x}_2$ with small error versus target image $x_2$ in pose $p_2$.

The proposed dense intrinsic appearance flow consists of two components, namely a flow map $F_{(x_1,x_2)}$ and a visibility map $V_{(x_1,x_2)}$ between image pair $(x_1,x_2)$ to jointly represent their pixel-wise correspondence in 3D space. In the following, we omit the subscript and brief them as $F$ and $V$ for simplicity. Note $F$ and $V$ have the same spatial dimensions as the target image $x_2$. Assume that $u_i'$ and $u_i$ are the 2D coordinates in images $x_1$ and $x_2$ that are projected from the same 3D body point $h_i$, $F$ and $V$ can be defined as:
\begin{equation}\label{eq_flow}
\begin{aligned}
f_i=&F(u_i)=u_i'-u_i, \\
v_i=&V(u_i)=visibility(h_i,x_1),
\end{aligned}
\end{equation}
where $visibility(h_i,x_1)$ is a function that indicates whether $h_i$ is invisible (due to self-occlusion or out of the image plane) in $x_1$. It outputs 3 discrete values (representing visible, invisible or background) which are color-coded in a visibility map $V$ (see an example in Fig.~\ref{fig:flow_module}).
\subsection{Overall Framework}
\label{sec:framework}
\begin{figure*}[t]
	\centering
	\includegraphics[width=0.95\linewidth]{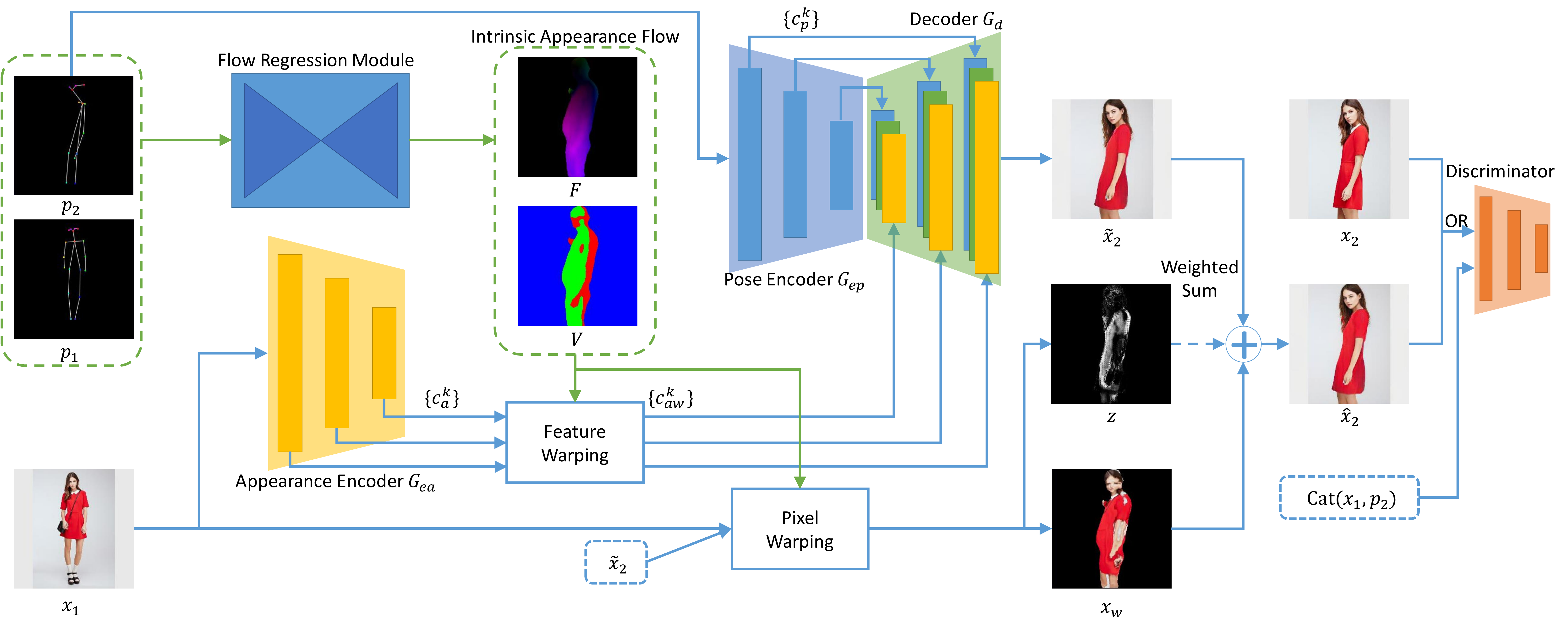}
	\vskip -0.25cm
	\caption{Overview of our human pose transfer framework. With the input image $x_1$, its extracted pose $p_1$, and the target pose $p_2$, the goal is to render a new image in pose $p_2$. Our flow regression module first generates the intrinsic appearance flow map $F$ and visibility map $V$, which are used to warp the encoded features $\{c^k_{a}\}$ from reference image $x_1$. Such warped features $\{c^k_{aw}\}$ and target pose features $\{c^k_{p}\}$ can then go through a decoder $G_d$ to produce an image $\widetilde{x}_2$. This result is further refined by a pixel warping module to generate the final result $\hat{x}_2$. Our training objectives include using the PatchGAN~\cite{pix2pix2017} to discriminate between $(x_1,p_2,x_2)$ and $(x_1,p_2,\hat{x}_2)$, as well as reconstruction and perceptual losses.}
	\label{fig:framework}
	\vskip -0.25cm
\end{figure*}

Figure.~\ref{fig:framework} illustrates our human pose transfer framework. Given the input image $x_1$ and its extracted pose $p_1$, together with the target pose $p_2$, the flow regression module first predicts from $(p_1,p_2)$ the intrinsic 3D appearance flow $F$ and visibility map $V$ by Eq.~\eqref{eq_flow}. 
Then we use the tuple $(x_1,p_2,F,V)$ for image generation. Note the input image $x_1$ and target pose $p_2$ are likely misaligned spatially, therefore if we want to directly concatenate and feed them into a single convolutional network to generate the target image, we can suffer from sub-optimal results. Part of the reason is that the convolutional layers (especially those low-level ones) in one single network may have limited receptive field to capture the large spatial displacements. Some unique network architecture is introduced in~\cite{ma2017pose} to address this.

Inspired by~\cite{siarohin2018deformable,zhao2018multi}, we choose to use a \textit{dual-path} U-Net~\cite{ronneberger2015u} to separately model the image and pose information. Concretely, an appearance encoder $G_{ea}$ and pose encoder $G_{ep}$ are employed to encode image $x_1$ and target pose $p_2$ into the feature pyramids $\{c_a^k\}$, $\{c_p^k\}$. Then a feature warping module is proposed to handle the spatial misalignment issue during pose transfer. This module warps the appearance features $c_a^k$ according to our generated flow map $F$. Meanwhile, some potentially missing pixels in target pose are also implicitly considered by including the visibility map $V$. Our feature warping function is defined as:
\begin{equation}\label{eq_feat_warp}
\begin{aligned}
c_{aw}^k=W_F(c_a^k, F, V),
\end{aligned}
\end{equation}
where $W_F$ is the warping operation detailed in Sec.~\ref{sec:feature_warping}, and $c^k_{aw}$ denotes the warped features at feature level $k$. Then we concatenate warped features $\{c_{aw}^k\}$ and target pose features $\{c_p^k\}$ hierarchically, which are fed to the image decoder $G_d$ through skip connections to generate the target image $\widetilde{x}_2$. Lastly, $\widetilde{x}_2$ is further enhanced by a pixel warping module (Sec.~\ref{sec:pixel_warping}) to obtain the finial output $\hat{x}_2$.

One of our training objectives is the adversarial loss. We adopt the PatchGAN~\cite{pix2pix2017} to score the realism of synthesized image patches. The input patches to the PatchGAN discriminator is either from $(x_1,p_2,x_2)$ or $(x_1,p_2,\hat{x}_2)$. We found the concatenation of $(x_1,p_2)$ provides good conditioning for GAN training.

\subsection{Flow Regression Module}
\label{sec:flow_module}
\begin{figure}[t]
	\centering
	\includegraphics[width=\linewidth]{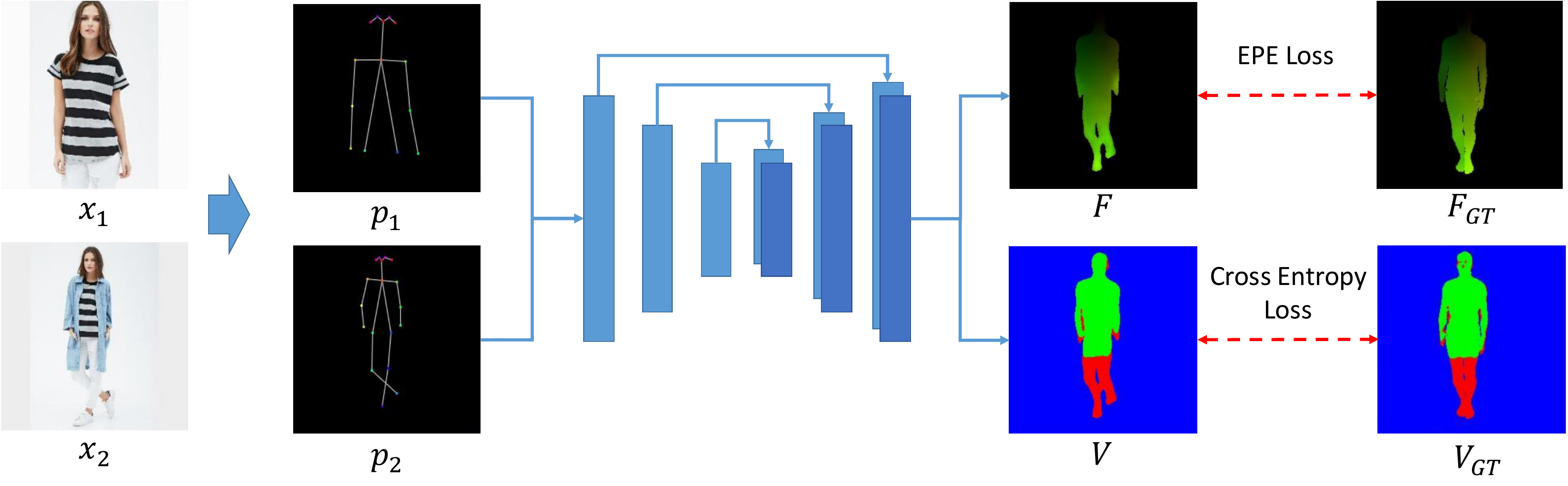}
	\vskip -0.25cm
	\caption{Our appearance flow regression module adopts a U-Net architecture to predict the intrinsic 3D appearance flow map $F$ and visibility map $V$ from the given pose pair $(p_1,p_2)$. This module is jointly trained with an End-Point-Error (EPE) loss on $F$ and a cross-entropy loss on $V$.}
	\label{fig:flow_module}
	\vskip -0.25cm
\end{figure}
\begin{figure}[t]
	\centering
	\includegraphics[width=\linewidth]{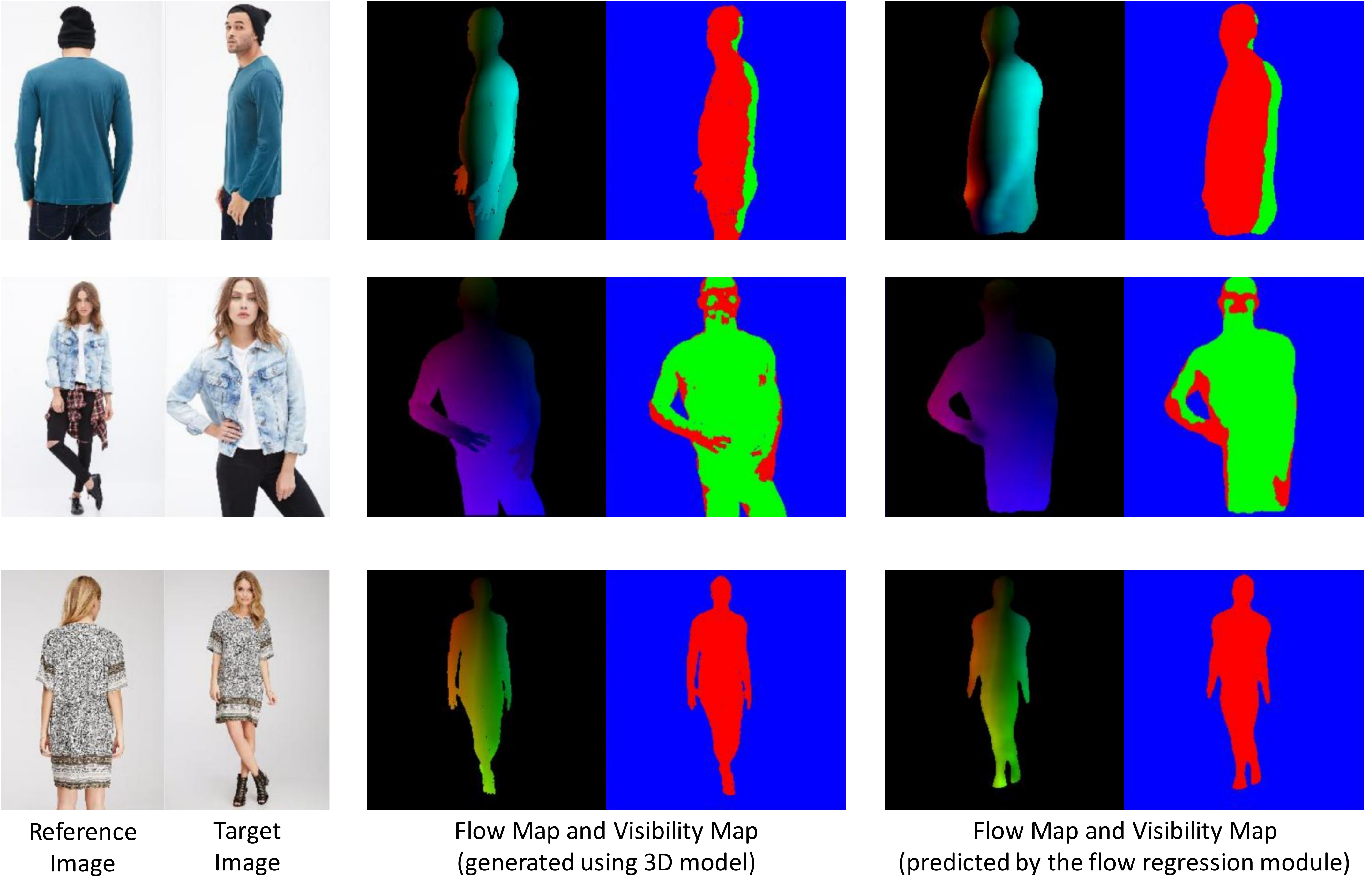}
	\caption{Example 3D appearance flow maps and visibility maps: generated ground-truth (middle) and prediction from our flow regression module (right). The ground-truth rendered from 3D model fitting has occasional errors,~\eg,~around the overlapping legs in the last row. While our flow regression module can correct the error by predicting from the given pose.}
	\label{fig:flow_dataset}
\end{figure}
Our key module for 3D appearance flow regression is shown in Fig.~\ref{fig:flow_module}. It is a feedforward CNN that predicts the required appearance flow map $F$ and visibility map $V$ from the pose pair $(p_1,p_2)$. This is similar to the optical flow prediction~\cite{dosovitskiy2015flownet,ilg2017flownet}, but differs in that our flow and visibility maps aim to encode 3D dense correspondences not 2D ones in optical flow. For accurate prediction of these two maps, we leverage a 3D human model to synthesize their ground-truth for training.

\noindent
\textbf{Ground-truth generation}. For this purpose, we randomly sample the same-person image pairs $(x_1, x_2)$ from the DeepFashion dataset~\cite{liu2016deepfashion}. We then fit a 3D human model~\cite{loper2015smpl} to both images, using the state-of-the-art method~\cite{kanazawa2018end}. The 3D model represents the human body as a mesh with 6,890 vertices and 13,766 faces. After 3D model fitting, we project them back to the 2D image plane using an image renderer~\cite{loper2014opendr}. As indicated by Eq.~\eqref{eq_flow}, for the projected 2D coordinate $u_j$ in image $x_2$, we can identify its exact belonging mesh face in 3D and hence compute the corresponding 2D coordinate $u_j'$ in image $x_1$ via barycentric transformation. The resulting flow vector is computed as $f_j=u_j'-u_j$. In addition, we can obtain the visibility of each mesh face and thus the entire visibility map $V$ from the image renderer. Fig.~\ref{fig:flow_dataset} (middle) shows some examples of the generated groundtruth flow map and visibility map. One by-product of the 2D image projection is that we can obtain the corresponding 2D pose from the image renderer~\cite{kanazawa2018end}. We denote such rendered pose as $\widetilde{p}$, and will elaborate its use next.

\noindent
\textbf{Network architecture and training}. Figure.~\ref{fig:flow_module} demonstrates how to train the 3D appearance flow regression module with a U-Net architecture. It takes a pose pair $(p_1,p_2)$ as input and is trained to simultaneously predict the flow map $F$ and visibility map $V$ under the end-point-error (EPE) loss and cross entropy loss, respectively.
We noticed that the 3D model fitting process will sometimes cause errors,~\eg,~when human legs are overlapped with each other, see Fig.~\ref{fig:flow_dataset} (middle, last row). In this case, the synthesized flow and visibility maps $\{F,V\}$ from image-based 3D fitting is not consistent with the groundtruth pose $(p_1,p_2)$ anymore. Hence it is erroneous to train the flow regression from $(p_1,p_2)$ to the un-matched $\{F,V\}$. Fortunately, as mentioned before, we have pose $\{\widetilde{p}_1, \widetilde{p}_2\}$ rendered from the 2D projection process that leads to the corresponding maps $\{F,V\}$. Therefore, we choose to perform regression from the rendered pose $(\widetilde{p}_1, \widetilde{p}_2)$ to $\{F,V\}$, rather than from the potentially un-matched ground-truth pose $(p_1,p_2)$. We found such trained regressor between the rendered pose-flow pair works surprisingly well even when the 3D model is not fitted perfectly. Once our appearance flow regression module finishes training, it is frozen during the training of the overall pose transfer framework. At test time, our flow regression module generalizes well to the given pose $p$.
%

\subsection{Flow-Guided Feature Warping}
\label{sec:feature_warping}
\begin{figure}[t]
	\centering
	\includegraphics[width=\linewidth]{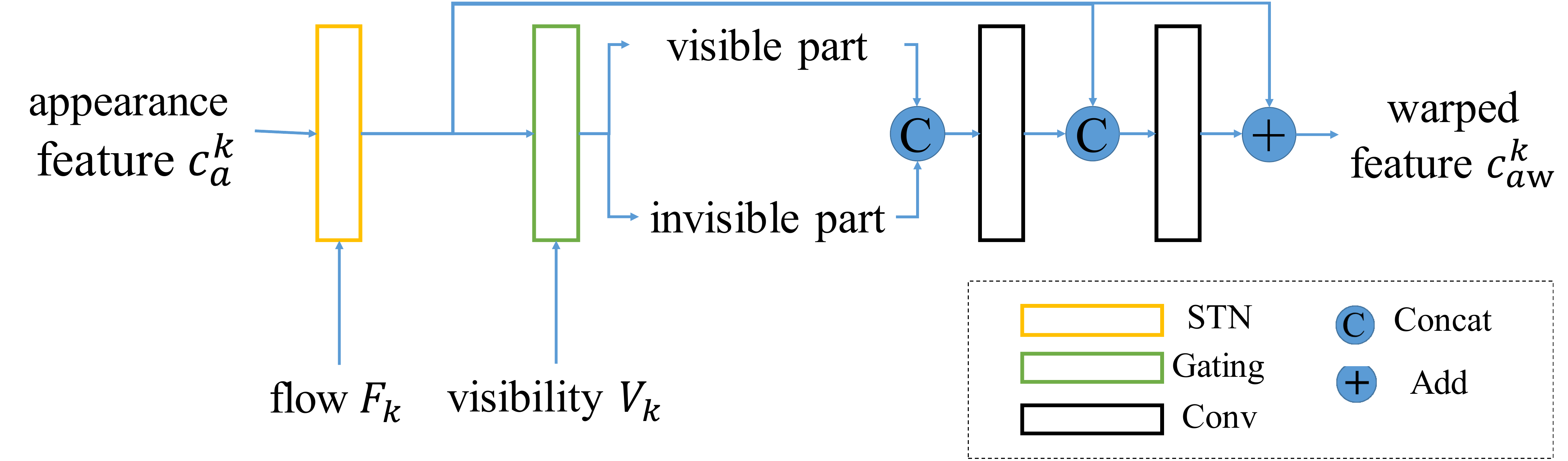}
	\caption{The architecture of feature warping module.}
	\label{fig:feature_warping_module}
	\vskip -0.25cm
\end{figure}
Recall that our 3D appearance flow and visibility maps are generated to align the reference image to the target pose and inpaint the invisible pixels therein. We achieve this by warping the input image features guided by our two maps. The architecture of our feature warping module is illustrated in Fig.~\ref{fig:feature_warping_module}. 
The inputs are the image features $c_a^k$ (at feature level $k$) and the flow and visibility maps $(F_k,V_k)$ resized to match the $c_a^k$ dimensions.
We first warp the input features $c_a^k$ by the flow map $F_k$ using a spatial transformer layer~\cite{jaderberg2015spatial}. The warped features are then fed into a spatial gating layer, which divides the feature maps into mutually exclusive regions according to the visibility map $V_k$. Here we do not simply filter out the invisible feature map pixels because they may still contain useful information, like clothing style or body shape. The gated feature maps are passed through two convolutional layers with residual path to get the final warped features $c_{aw}^k$. Our feature warping module is differentiable allowing for end-to-end training.
\subsection{Pixel Warping}
As shown in Fig.~\ref{fig:framework}, given the warped features $\{c_{aw}^k\}$, we concatenate them with the target pose features $\{c_p^k\}$ hierarchically. They are both fed to the image decoder $G_d$ through skip connections to render the target image $\widetilde{x}_2$. In our experiments, we found high frequency details are sometimes lost in $\widetilde{x}_2$, indicating the inefficiency of image warping only at feature level. To this end, we propose to further enhance $\widetilde{x}_2$ at pixel level. Similarly, a pixel warping module is adopted to warp the pixels in input image $x_1$ to the target pose using our 3D appearance flow.

Specifically, we warp $x_1$ according to the full resolution flow map $F$ to get the warped image $x_w$. Note $x_w$ contains the required image details from input $x_1$, but may be distorted because of the coarse flow map and body occlusions. Therefore, we train another U-Net to weigh between the warped output $x_w$ and $\widetilde{x}_2$ at pixel- and feature-level respectively. This weighting network takes $x_w$, $\widetilde{x}_2$, $F$ and $V$ as inputs and outputs a soft weighting map $z$ with the same resolution of $x_w$ and $x_2$. The map $z$ is normalized to the range of $(0,1)$ with sigmoid function. Then the final output $x^*_2$ is computed as a weighted sum of $x_w$ and $\widetilde{x}_2$ as:
\begin{equation}\label{eq_blend}
\begin{aligned}
\hat{x}_2 = z \cdot x_w + (1-z) \cdot \widetilde{x}_2.
\end{aligned}
\end{equation}

\begin{figure}[t]
	\centering
	\includegraphics[width=\linewidth]{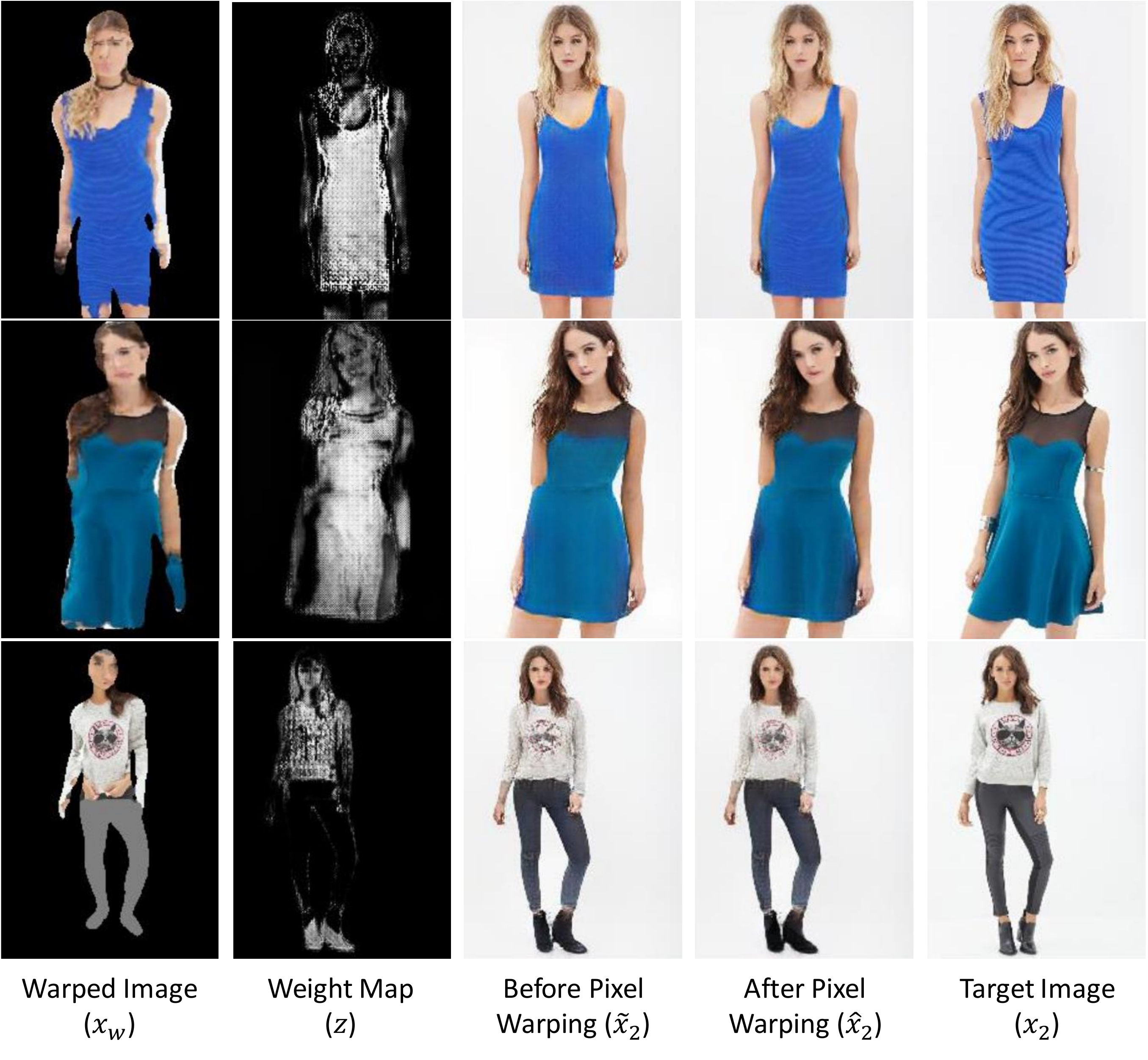}
	\vskip -0.25cm
	\caption{Pixel warping examples. From left to right: the pixel warped image $x_w$, weighting map $z$, feature-warped image $\widetilde{x}_2$, final image $\hat{x}_2$ fused with pixel warping, and the ground-truth target image $x_2$.}
	\label{fig:pixel_warping}
	\vskip -0.25cm
\end{figure}

Figure.~\ref{fig:pixel_warping} validates the effect of pixel warping. We can see that pixel warping is indeed able to add some high-frequency details that can not be recovered well by our feature warping results. The added details are simply copied from reference image using our intrinsic appearance flow.

\label{sec:pixel_warping}

\subsection{Loss Functions}
\label{sec:loss_functions}
The goal of our model is to achieve accurate human pose transfer to an arbitrary pose, generating a photorealistic pose-transferred image. This task is challenging due to the large non-rigid deformation during pose transfer and the complex details in human images. Previous works on conditional image generation~\cite{pix2pix2017,wang2018high} and human pose transfer~\cite{ma2017pose,neverova2018dense,siarohin2018deformable} utilize multiple loss functions to jointly supervise the training process. In this work we similarly use a combination of three loss functions, namely an adversarial loss $\mathcal{L}_{adv}$, an L1 reconstruction loss $\mathcal{L}_{L1}$, and a perceptual loss $\mathcal{L}_{perceptual}$. They are detailed as follows.

\noindent
\textbf{Adversarial loss.} We adopt a vanilla GAN loss in the conditional setting in our task, which is defined as:
\begin{equation}\label{eq_loss_gan}
\begin{aligned}
\mathcal{L}_{adv}(G,D) = &E_{x_1,x_2}[logD(x_2|x_1,p_2)] \\+ &E_{x_1,x_2}[log(1-D(G(x_1,p_2)|x_1,p_2))].
\end{aligned}
\end{equation}

\noindent
\textbf{L1 loss.} Previous work~\cite{pix2pix2017} shows L1 loss can stabilize the training process when a target groundtruth is available. Therefore we also enforce an L1 constraint between the generated image and the target image as:
\begin{equation}\label{eq_loss_l1}
\begin{aligned}
\mathcal{L}_{L1}(G) = ||\hat{x}_2 - x_2||_1.
\end{aligned}
\end{equation}

\noindent
\textbf{Perceptual loss.} The work in~\cite{johnson2016perceptual} shows that penalizing L2-distance between feature maps extracted from two images by a pretrained CNN could encourage image structure similarity. We adopt a VGG19 network~\cite{simonyan2014very} pretrained on ImageNet~\cite{russakovsky2015imagenet} as the feature extractor, and use multi-level feature maps $\phi_j$ to compute perceptual loss as:
\begin{equation}\label{eq_loss_perceptual}
\begin{aligned}
\mathcal{L}_{perceptual}(G) = \sum_{j=1}^{N}||\phi_j(\hat{x}_2)-\phi_j(x_2)||^2_2.
\end{aligned}
\end{equation}
Our final loss function for image generation is a weighted sum of above terms:
\begin{equation}\label{eq_loss_total}
\begin{aligned}
\mathcal{L}(G) = \lambda_1\mathcal{L}_{adv}+\lambda_2\mathcal{L}_{L1}+\lambda_3\mathcal{L}_{perceptual}.
\end{aligned}
\end{equation}

\section{Experiments}
\label{sec:experiments}
\subsection{Dataset and Implementation Details}
\label{sec:dataset_and_implementation}
\noindent
\textbf{Dataset}. We evaluate our method on DeepFashion dataset (In-shop Clothes Retrieval Benchmark)~\cite{liu2016deepfashion}, which contains 52,712 in-shop clothes images and 200,000 cross-pose/scale pairs. The images have a resolution of $256\times256$ pixels. Following the setting in~\cite{siarohin2018deformable}, we select 89,262 pairs for training and 12,000 pairs for testing. We perform additional experiments on Market-1501 dataset~\cite{zheng2015scalable} and show results in the supplementary material.

\noindent
\textbf{Network architecture}. Our generator uses a U-Net architecture of $N=7$ levels. At each feature level, the encoder has two cascaded residual blocks~\cite{he2016identity} followed by a stride-2 convolution layer for downsampling, while the decoder has a symmetric structure of an upsampling layer followed by two residual blocks. The upsampling layer is implemented as a convolutional layer followed by pixel shuffling operation~\cite{shi2016real}. There are skip connections between the corresponding residual blocks in the encoder and decoder, and batch normalization~\cite{ioffe2015batch} is used after each convolutional layer (except the last one). Our discriminator uses the PatchGAN~\cite{pix2pix2017} network with a patch size of $70\times70$ pixels.

\noindent
\textbf{Training}. We use the Adam optimizer~\cite{kingma2014adam} ($\beta_1=0.5$,$\beta_2=0.999$) in all experiments. We adopt a batch size of 8 and a learning rate of 2e-4 (except for the discriminator which uses learning rate 2e-5). In our experiments we noticed that optimizing the image generator and the pixel warping module separately yields better performance. Therefore we first train the image generator for 10 epochs and freeze it afterwards. Then we add the pixel warping module into the framework and train the full model for another 2 epochs. To stabilize the training, $\mathcal{L}_{GAN}$ is not used in the first 5 epochs.
\subsection{Evaluation Metrics}
\label{sec:metrics}
Previous works use Structure Similarity (SSIM)~\cite{wang2004image} and Inception Score (IS)~\cite{salimans2016improved} to evaluate the quality of generated images. We report these metrics too in our experiments. However, SSIM is noticed to favor blurry images which are less photorealistic~\cite{ma2017pose}. While IS computed using a classifier trained on ImageNet~\cite{russakovsky2015imagenet} is not suitable in the scenario where the images have a different distribution than ImageNet images. For these reasons, we introduce two complementary metrics described below.

\noindent
\textbf{Fashion inception score}. Following the definition in~\cite{salimans2016improved}, we calculate the inception score using a fashion item classifier, which we refer as FashionIS. Specifically, we finetune an Inception Model~\cite{szegedy2016rethinking} on clothing type classification task on~\cite{liu2016deepfashion}, which has no domain gap to the images in our human pose transfer experiments. We argue that FashionIS can better evaluate the image quality in our experiments compared to the original IS.

\noindent
\textbf{Clothing attribute retaining rate}. The human pose transfer model should be able to preserve the appearance details in the reference image, like the clothing attributes like color, texture, fabric and style. To evaluate the model performance from this aspect, we train a clothing attribute recognition model on DeepFashion~\cite{liu2016deepfashion} to recognize clothing attributes from the generated images. Since the groundtruth attribute label of the test image is available, we directly use the top-k recall rate as the metric, denoted as AttrRec-k.
\subsection{Quantitative Results}
\label{sec:quantitative_results}
\begin{table}[t]
	\centering
	\caption{Comparison against previous works on DeepFashion dataset. $\dagger$ indicates the model is unsupervised (no image pairs used in training). $*$ indicates the results are obtained using different data splits, thus cannot be directly compared to ours.}
	
	\label{table:compare}
	\small{
		\begin{tabular}{c|c c c c c}
			\hline
			\multirow{2}{*}{Model} & \multirow{2}{*}{SSIM} &\multirow{2}{*}{IS} & \multirow{2}{*}{FashionIS} & \multicolumn{2}{c}{AttrRec-k(\%)}\\
			\hhline{~~~~--}
			&&&& k=5 & k=20\\
			\hline
			UPIS~\cite{pumarola2018unsupervised}$\dagger\ast$ & 0.747 & 2.97 & - & - & -\\
			DPT~\cite{neverova2018dense}$\ast$ & 0.796 & 3.71 & - & - & -\\
			\hline
			DPIG~\cite{ma2018disentangled}$\dagger$ & 0.614 & 3.228 & - & - & -\\
			VUnet~\cite{esser2018variational}$\dagger$ & \textbf{0.786} & 3.087 & - & - & -\\
			PG2~\cite{ma2017pose} & 0.762 & 3.090 & 2.639 & 13.560 & 30.193\\
			DSC~\cite{siarohin2018deformable} & 0.756 & \textbf{3.439} & 3.804 & 19.017 & 43.812\\
			Ours & 0.778 & 3.338 & \textbf{4.898} & \textbf{21.065} & \textbf{49.044} \\
			\hline
			Real Image & 1.000 & 3.962 & 6.518 & 24.780 & 61.626 \\
			\hline
		\end{tabular}
	}
	\vskip -0.2cm
\end{table}
We compare our proposed method against recent works in Table.~\ref{table:compare}. For SSIM and IS we directly use the results reported in the original papers. We calculate their FashionIS and AttRec-k results using the images generated by the publicly released codes and models. Note that the data splits used in ~\cite{neverova2018dense,pumarola2018unsupervised} are different from our setting, thus these results are not directly comparable. The results show that our proposed method outperforms others in terms of both FashionIS and AttrRec-k metrics by a significant margin. This proves that our method can generate more realistic images with better preserved details. In terms of SSIM and IS, we also achieve compelling results compared to the state-of-the-art methods.
\subsection{Qualitative Results}
\label{sec:qualitative_results}
\begin{figure}[t]
	\centering
	\includegraphics[width=\linewidth]{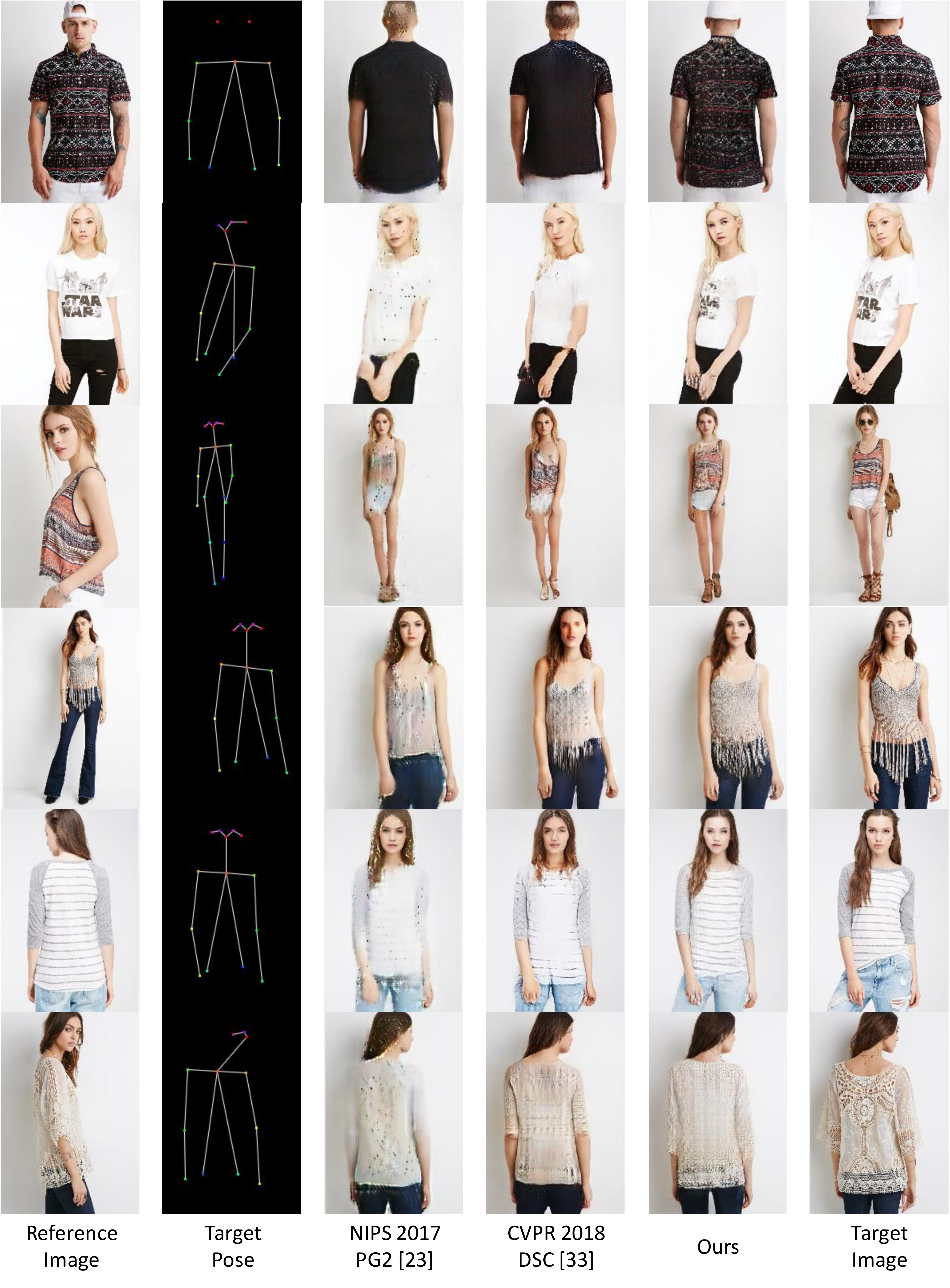}
	\caption{Qualitative comparison between our method and previous works.}
	\label{fig:qualitative}
	\vskip -0.2cm
\end{figure}
We further visualize some qualitative results in Fig.~\ref{fig:qualitative} to show the effectiveness of our proposed method. Because of the introduced 3D intrinsic appearance flow and visibility map, the large spatial displacements and deformations are successfully recovered by our method during pose transfer. We can see that our model generates realistic human image in arbitrary poses and is able to restore detailed appearance attributes like clothing textures.

\subsection{User Study}
\label{sec:user_study}
\begin{table}[t]
	\centering
	\caption{User study (\%) on DeepFashion. R2G indicates the percentage of real images rated as fake, and G2R means the opposite. 'Judged as better' indicates the wining percentage in the comparison test.}
	\label{table:user_study}
	\small{
		\begin{tabular}{c|c c|c}
			\hline
			Model & R2G & G2R & Judged as better\\
			\hline
			DSC~\cite{siarohin2018deformable} & 9.55 & 9.24 & 9.47\\
			Ours & 10.01 & 31.71 & 90.53\\
			\hline			
		\end{tabular}
	}
	\vskip -0.2cm
\end{table}
We conduct a user study with 30 users to compare the visual results from our method and the state-of-the-art baseline~\cite{siarohin2018deformable}. The user study consists of two tests. The first one is a "real or fake" test, following the protocol in~\cite{ma2017pose,siarohin2018deformable}. For each method, we show the user 55 real images and 55 fake images in an random order. Each image is shown for 1 second and user will determine whether it is real or fake. The first 10 images are for practice and are ignored when computing results. The second one is a comparison test, in which we show the user 55 image pairs, generated by our method and baseline respectively with the same reference image and target pose, and the user is asked to pick one image with better quality from each pair. The reference image is also shown to make the user aware of the groundtruth appearance. Similar to the first test, the first 5 pairs are for practice. All samples in user study is randomly selected from our test set and shown with full resolution. The results in Table.~\ref{table:user_study} show that our method generates images with consistently better quality than the baseline, which are confused with real images more often by human judeges.

\subsection{Ablation Study}
\label{sec:ablation}
\begin{table}[b]
	\centering
	\vskip -0.4cm
	\caption{Ablation study.}
	\vskip -0.1cm
	\label{table:ablation}
	\small{
		\begin{tabular}{c|c c @{ }c@{ } c c}
			\hline
			\multirow{2}{*}{Model} & \multirow{2}{*}{SSIM} &\multirow{2}{*}{IS} & \multirow{2}{*}{FashionIS} & \multicolumn{2}{c}{AttrRec-k(\%)}\\
			\hhline{~~~~--}
			&&&& k=5 & k=20\\
			\hline
			w/o. dual encoder & 0.780 & 3.173 & 3.927 & 19.085 & 43.377 \\
			w/o. flow & \textbf{0.783} & 3.319 & 4.119 & 19.716 & 44.656 \\
			w/o. visibility & 0.778 & 3.260 & 4.491 & 20.297 & 46.591 \\
			w/o. pixel warping & 0.776 & 3.281 & 4.800 & 20.942 & 48.391\\
			Full & 0.778 & \textbf{3.338} & \textbf{4.898} & \textbf{21.065} & \textbf{49.044} \\
			\hline
		\end{tabular}
	\vskip -0.2cm
	}
\end{table}

In this section we perform ablation study to further analyze the impact of each component in our model. We first describe the variants obtained by incrementally removing components form the full framework. All variants are trained using the same protocol described in Sec.~\ref{sec:dataset_and_implementation}. 

\noindent
\textbf{w/o. dual encoder}. This is similar to PG2~\cite{ma2017pose} that has a U-Net architecture with single encoder and no flow regression module. $x_1$ and $p_2$ are concatenated before being fed into the model.

\noindent
\textbf{w/o. flow}. This model has a dual-path U-Net architecture but without feature warping module. Appearance features $\{c_a^k\}$ and pose features $\{c_p^k\}$ are directly concatenated at corresponding level before sent into the decoder.

\noindent
\textbf{w/o. visibility}. This model adopts dual-path U-Net generator with a simplified feature warping module, where the gating layer and the first convolution layer in Fig.~\ref{fig:feature_warping_module} are replaced with a normal residual block that is unaware of the visibility map $V$.

\noindent
\textbf{w/o. pixel warping}. This model uses the full generators in Fig.~\ref{fig:framework} without pixel warping module.

\noindent
\textbf{Full}. This is the full framework as shown in Fig.~\ref{fig:framework}.

\begin{figure}[t]
	\centering
	\includegraphics[width=\linewidth]{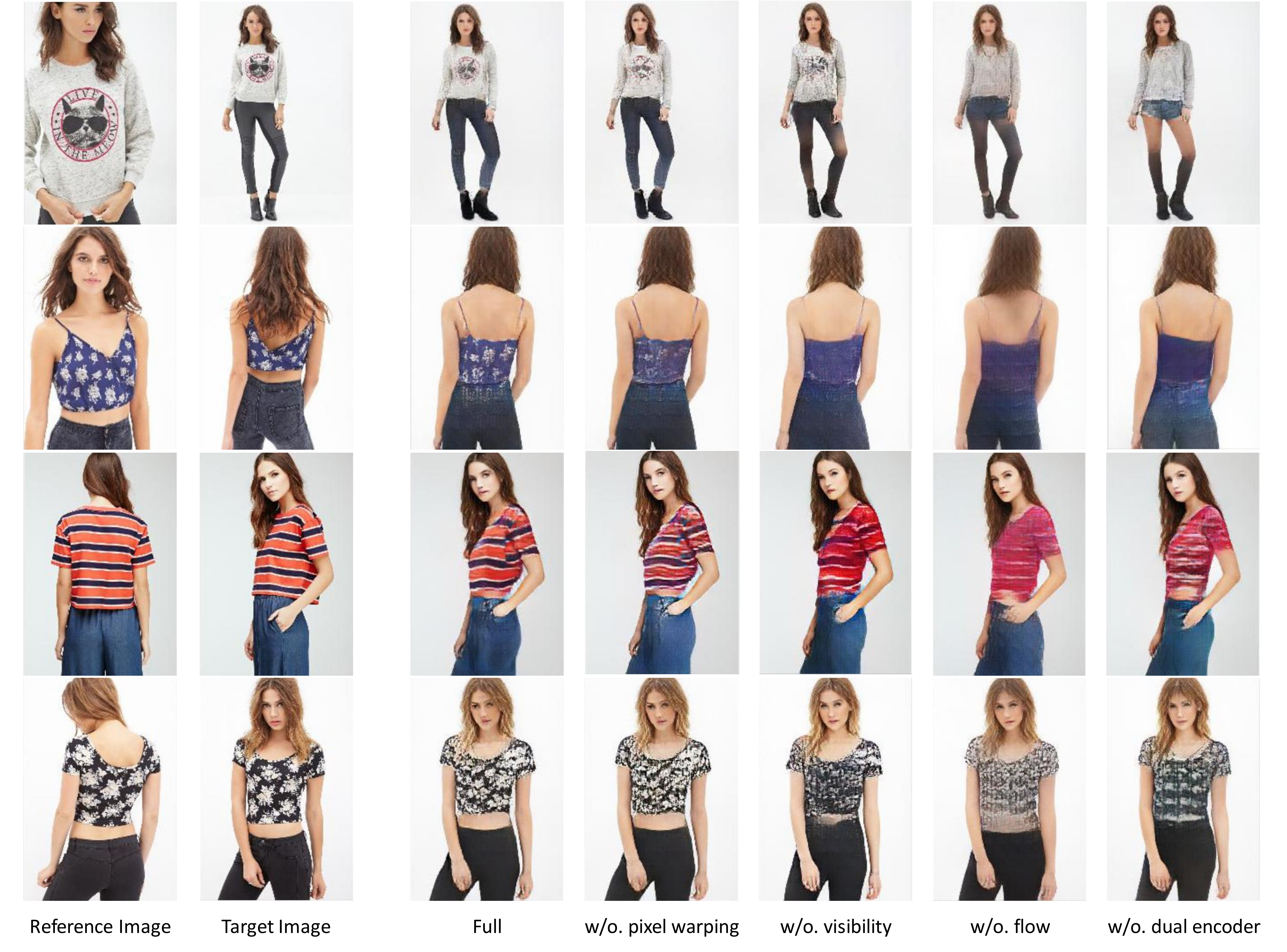}
	\caption{Visualization of ablation study}
	\label{fig:ablation}
	\vskip -0.5cm
\end{figure}

Table.~\ref{table:ablation} and Fig.~\ref{fig:ablation} show the quantitative and qualitative results of the ablation study. We can observe that all models perform well on generating correct body poses, realistic faces and plausible color style, which yield high SSIM scores. However, our proposed flow guided feature warping significantly improves the capability of preserving detailed appearance attributes like clothing layout and complex textures, which also leads a large increase of FashionIS and AttrRec-k. The pixel warping module further helps to handle some special clothing patterns that are not well reconstructed by the convolutional generator.

\section{Conclusion}
\label{sec:conclusion}

In this paper we propose a new human pose transfer method with implicit reasoning about 3D geometry of human body. We generate the intrinsic appearance flow map and visibility map leveraging the 3D human model, so as to learn how to move pixels and hallucinate invisible ones in the target pose. A feedforward neural network is trained to rapidly predict both maps, which are used to warp and gate image features respectively for high-fidelity image generation. Both qualitative and quantitative results on the DeepFashion dataset show that our method is able to synthesize human images in arbitrary pose with realistic details and preserved attributes. Our approach also significantly outperforms existing pose- or keypoint-based image generators and other alternatives.

\noindent \textbf{Acknowledgement}: This work is supported by SenseTime Group Limited, the General Research Fund sponsored by the Research Grants Council of the Hong Kong SAR (CUHK 14241716, 14224316. 14209217), and Singapore MOE AcRF Tier 1 (M4012082.020).

{\small
\bibliographystyle{ieee_fullname}
\bibliography{ms}

\begin{thebibliography}{10}\itemsep=-1pt

\bibitem{balakrishnan2018synthesizing}
Guha Balakrishnan, Amy Zhao, Adrian~V Dalca, Fredo Durand, and John Guttag.
\newblock Synthesizing images of humans in unseen poses.
\newblock In {\em CVPR}, 2018.

\bibitem{cao2017realtime}
Zhe Cao, Tomas Simon, Shih-En Wei, and Yaser Sheikh.
\newblock Realtime multi-person 2d pose estimation using part affinity fields.
\newblock In {\em CVPR}, 2017.

\bibitem{chan2018everybody}
Caroline Chan, Shiry Ginosar, Tinghui Zhou, and Alexei~A Efros.
\newblock Everybody dance now.
\newblock {\em arXiv preprint arXiv:1808.07371}, 2018.

\bibitem{dosovitskiy2015flownet}
Alexey Dosovitskiy, Philipp Fischer, Eddy Ilg, Philip Hausser, Caner Hazirbas,
  Vladimir Golkov, Patrick Van Der~Smagt, Daniel Cremers, and Thomas Brox.
\newblock Flownet: Learning optical flow with convolutional networks.
\newblock In {\em ICCV}, 2015.

\bibitem{esser2018variational}
Patrick Esser, Ekaterina Sutter, and Bj{\"o}rn Ommer.
\newblock A variational u-net for conditional appearance and shape generation.
\newblock In {\em CVPR}, 2018.

\bibitem{goodfellow2014generative}
Ian Goodfellow, Jean Pouget-Abadie, Mehdi Mirza, Bing Xu, David Warde-Farley,
  Sherjil Ozair, Aaron Courville, and Yoshua Bengio.
\newblock Generative adversarial nets.
\newblock In {\em NIPS}, 2014.

\bibitem{guler2018densepose}
Riza~Alp G{\"{u}}ler, Natalia Neverova, and Iasonas Kokkinos.
\newblock Densepose: Dense human pose estimation in the wild.
\newblock In {\em CVPR}, 2018.

\bibitem{he2016identity}
Kaiming He, Xiangyu Zhang, Shaoqing Ren, and Jian Sun.
\newblock Identity mappings in deep residual networks.
\newblock In {\em ECCV}, 2016.

\bibitem{horn1981determining}
Berthold~KP Horn and Brian~G Schunck.
\newblock Determining optical flow.
\newblock {\em Artificial intelligence}, 17(1-3):185--203, 1981.

\bibitem{ilg2017flownet}
Eddy Ilg, Nikolaus Mayer, Tonmoy Saikia, Margret Keuper, Alexey Dosovitskiy,
  and Thomas Brox.
\newblock Flownet 2.0: Evolution of optical flow estimation with deep networks.
\newblock In {\em CVPR}, 2017.

\bibitem{ioffe2015batch}
Sergey Ioffe and Christian Szegedy.
\newblock Batch normalization: Accelerating deep network training by reducing
  internal covariate shift.
\newblock {\em arXiv preprint arXiv:1502.03167}, 2015.

\bibitem{pix2pix2017}
Phillip Isola, Jun-Yan Zhu, Tinghui Zhou, and Alexei~A Efros.
\newblock Image-to-image translation with conditional adversarial networks.
\newblock In {\em CVPR}, 2017.

\bibitem{jaderberg2015spatial}
Max Jaderberg, Karen Simonyan, Andrew Zisserman, et~al.
\newblock Spatial transformer networks.
\newblock In {\em NIPS}, 2015.

\bibitem{johnson2016perceptual}
Justin Johnson, Alexandre Alahi, and Li Fei-Fei.
\newblock Perceptual losses for real-time style transfer and super-resolution.
\newblock In {\em ECCV}, 2016.

\bibitem{kanazawa2018end}
Angjoo Kanazawa, Michael~J Black, David~W Jacobs, and Jitendra Malik.
\newblock End-to-end recovery of human shape and pose.
\newblock In {\em CVPR}, 2018.

\bibitem{kingma2014adam}
Diederik~P Kingma and Jimmy Ba.
\newblock Adam: A method for stochastic optimization.
\newblock {\em arXiv preprint arXiv:1412.6980}, 2014.

\bibitem{kingma2013auto}
Diederik~P Kingma and Max Welling.
\newblock Auto-encoding variational bayes.
\newblock {\em arXiv preprint arXiv:1312.6114}, 2013.

\bibitem{lassner2017generative}
Christoph Lassner, Gerard Pons-Moll, and Peter~V Gehler.
\newblock A generative model of people in clothing.
\newblock In {\em ICCV}, 2017.

\bibitem{ledig2017photo}
Christian Ledig, Lucas Theis, Ferenc Husz{\'a}r, Jose Caballero, Andrew
  Cunningham, Alejandro Acosta, Andrew~P Aitken, Alykhan Tejani, Johannes Totz,
  Zehan Wang, et~al.
\newblock Photo-realistic single image super-resolution using a generative
  adversarial network.
\newblock In {\em CVPR}, 2017.

\bibitem{liu2016deepfashion}
Ziwei Liu, Ping Luo, Shi Qiu, Xiaogang Wang, and Xiaoou Tang.
\newblock Deepfashion: Powering robust clothes recognition and retrieval with
  rich annotations.
\newblock In {\em CVPR}, 2016.

\bibitem{loper2015smpl}
Matthew Loper, Naureen Mahmood, Javier Romero, Gerard Pons-Moll, and Michael~J
  Black.
\newblock Smpl: A skinned multi-person linear model.
\newblock {\em ACM TOG}, 34(6):248, 2015.

\bibitem{loper2014opendr}
Matthew~M Loper and Michael~J Black.
\newblock Opendr: An approximate differentiable renderer.
\newblock In {\em ECCV}, 2014.

\bibitem{ma2017pose}
Liqian Ma, Xu Jia, Qianru Sun, Bernt Schiele, Tinne Tuytelaars, and Luc
  Van~Gool.
\newblock Pose guided person image generation.
\newblock In {\em NIPS}, 2017.

\bibitem{ma2018disentangled}
Liqian Ma, Qianru Sun, Stamatios Georgoulis, Luc Van~Gool, Bernt Schiele, and
  Mario Fritz.
\newblock Disentangled person image generation.
\newblock In {\em CVPR}, 2018.

\bibitem{mirza2014conditional}
Mehdi Mirza and Simon Osindero.
\newblock Conditional generative adversarial nets.
\newblock {\em arXiv preprint arXiv:1411.1784}, 2014.

\bibitem{neverova2018dense}
Natalia Neverova, R{\i}za~Alp G{\"u}ler, and Iasonas Kokkinos.
\newblock Dense pose transfer.
\newblock In {\em ECCV}, 2018.

\bibitem{pumarola2018unsupervised}
Albert Pumarola, Antonio Agudo, Alberto Sanfeliu, and Francesc Moreno-Noguer.
\newblock Unsupervised person image synthesis in arbitrary poses.
\newblock In {\em CVPR}, 2018.

\bibitem{raj2018swapnet}
Amit Raj, Patsorn Sangkloy, Huiwen Chang, James Hays, Duygu Ceylan, and Jingwan
  Lu.
\newblock Swapnet: Image based garment transfer.
\newblock In {\em ECCV}, 2018.

\bibitem{ronneberger2015u}
Olaf Ronneberger, Philipp Fischer, and Thomas Brox.
\newblock U-net: Convolutional networks for biomedical image segmentation.
\newblock In {\em MICCAI}, 2015.

\bibitem{russakovsky2015imagenet}
Olga Russakovsky, Jia Deng, Hao Su, Jonathan Krause, Sanjeev Satheesh, Sean Ma,
  Zhiheng Huang, Andrej Karpathy, Aditya Khosla, Michael Bernstein, et~al.
\newblock Imagenet large scale visual recognition challenge.
\newblock {\em IJCV}, 115(3):211--252, 2015.

\bibitem{salimans2016improved}
Tim Salimans, Ian Goodfellow, Wojciech Zaremba, Vicki Cheung, Alec Radford, and
  Xi Chen.
\newblock Improved techniques for training gans.
\newblock In {\em NIPS}, 2016.

\bibitem{shi2016real}
Wenzhe Shi, Jose Caballero, Ferenc Husz{\'a}r, Johannes Totz, Andrew~P Aitken,
  Rob Bishop, Daniel Rueckert, and Zehan Wang.
\newblock Real-time single image and video super-resolution using an efficient
  sub-pixel convolutional neural network.
\newblock In {\em CVPR}, 2016.

\bibitem{zhu2017be}
Raquel Urtasun Dahua Lin Chen Change~Loy Shizhan~Zhu, Sanja~Fidler.
\newblock Be your own prada: Fashion synthesis with structural coherence.
\newblock In {\em ICCV}, 2017.

\bibitem{siarohin2018deformable}
Aliaksandr Siarohin, Enver Sangineto, St{\'e}phane Lathuili{\`e}re, and Nicu
  Sebe.
\newblock Deformable gans for pose-based human image generation.
\newblock In {\em CVPR}, 2018.

\bibitem{simonyan2014two}
Karen Simonyan and Andrew Zisserman.
\newblock Two-stream convolutional networks for action recognition in videos.
\newblock In {\em NIPS}, 2014.

\bibitem{simonyan2014very}
Karen Simonyan and Andrew Zisserman.
\newblock Very deep convolutional networks for large-scale image recognition.
\newblock {\em arXiv preprint arXiv:1409.1556}, 2014.

\bibitem{szegedy2016rethinking}
Christian Szegedy, Vincent Vanhoucke, Sergey Ioffe, Jon Shlens, and Zbigniew
  Wojna.
\newblock Rethinking the inception architecture for computer vision.
\newblock In {\em CVPR}, 2016.

\bibitem{wang2018high}
Ting-Chun Wang, Ming-Yu Liu, Jun-Yan Zhu, Andrew Tao, Jan Kautz, and Bryan
  Catanzaro.
\newblock High-resolution image synthesis and semantic manipulation with
  conditional gans.
\newblock In {\em CVPR}, 2018.

\bibitem{wang2018recovering}
Xintao Wang, Ke Yu, Chao Dong, and Chen Change~Loy.
\newblock Recovering realistic texture in image super-resolution by deep
  spatial feature transform.
\newblock In {\em CVPR}, 2018.

\bibitem{wang2018esrgan}
Xintao Wang, Ke Yu, Shixiang Wu, Jinjin Gu, Yihao Liu, Chao Dong, Yu Qiao, and
  Chen~Change Loy.
\newblock Esrgan: Enhanced super-resolution generative adversarial networks.
\newblock In {\em ECCV}, 2018.

\bibitem{wang2004image}
Zhou Wang, Alan~C Bovik, Hamid~R Sheikh, and Eero~P Simoncelli.
\newblock Image quality assessment: from error visibility to structural
  similarity.
\newblock {\em IEEE transactions on image processing}, 13(4):600--612, 2004.

\bibitem{xian2016texturegan}
Varun Agrawal Amit Raj Jingwan Lu Chen Fang Fisher Yu James~Hays Wenqi~Xian,
  Patsorn~Sangkloy.
\newblock Texturegan: Controlling deep image synthesis with texture patches.
\newblock {\em CVPR}, 2018.

\bibitem{zanfir2018monocular}
Andrei Zanfir, Elisabeta Marinoiu, and Cristian Sminchisescu.
\newblock Monocular 3d pose and shape estimation of multiple people in natural
  scenes--the importance of multiple scene constraints.
\newblock In {\em CVPR}, 2018.

\bibitem{zanfir2018human}
Mihai Zanfir, Alin-Ionut Popa, Andrei Zanfir, and Cristian Sminchisescu.
\newblock Human appearance transfer.
\newblock In {\em CVPR}, 2018.

\bibitem{zhang2017stackgan++}
Han Zhang, Tao Xu, Hongsheng Li, Shaoting Zhang, Xiaogang Wang, Xiaolei Huang,
  and Dimitris Metaxas.
\newblock Stackgan++: Realistic image synthesis with stacked generative
  adversarial networks.
\newblock {\em arXiv preprint arXiv:1710.10916}, 2017.

\bibitem{zhang2017stackgan}
Han Zhang, Tao Xu, Hongsheng Li, Shaoting Zhang, Xiaogang Wang, Xiaolei Huang,
  and Dimitris~N Metaxas.
\newblock Stackgan: Text to photo-realistic image synthesis with stacked
  generative adversarial networks.
\newblock In {\em ICCV}, 2017.

\bibitem{zhao2018multi}
Bo Zhao, Xiao Wu, Zhi-Qi Cheng, Hao Liu, Zequn Jie, and Jiashi Feng.
\newblock Multi-view image generation from a single-view.
\newblock In {\em ACMMM}, 2018.

\bibitem{zheng2015scalable}
Liang Zheng, Liyue Shen, Lu Tian, Shengjin Wang, Jingdong Wang, and Qi Tian.
\newblock Scalable person re-identification: A benchmark.
\newblock In {\em ICCV}, 2015.

\bibitem{zhou2016view}
Tinghui Zhou, Shubham Tulsiani, Weilun Sun, Jitendra Malik, and Alexei~A Efros.
\newblock View synthesis by appearance flow.
\newblock In {\em ECCV}, 2016.

\end{thebibliography}


\begin{thebibliography}{1}\itemsep=-1pt

\bibitem{liu2016deepfashion}
Ziwei Liu, Ping Luo, Shi Qiu, Xiaogang Wang, and Xiaoou Tang.
\newblock Deepfashion: Powering robust clothes recognition and retrieval with
  rich annotations.
\newblock In {\em CVPR}, 2016.

\bibitem{ma2017pose}
Liqian Ma, Xu Jia, Qianru Sun, Bernt Schiele, Tinne Tuytelaars, and Luc
  Van~Gool.
\newblock Pose guided person image generation.
\newblock In {\em NIPS}, 2017.

\bibitem{siarohin2018deformable}
Aliaksandr Siarohin, Enver Sangineto, St{\'e}phane Lathuili{\`e}re, and Nicu
  Sebe.
\newblock Deformable gans for pose-based human image generation.
\newblock In {\em CVPR}, 2018.

\bibitem{zheng2015scalable}
Liang Zheng, Liyue Shen, Lu Tian, Shengjin Wang, Jingdong Wang, and Qi Tian.
\newblock Scalable person re-identification: A benchmark.
\newblock In {\em ICCV}, 2015.

\end{thebibliography}
}

\end{document}


\title{Dense Intrinsic Appearance Flow for Human Pose Transfer\newline Supplementary Material}

\author{
	\makebox[\linewidth][c]{Yining Li\Mark{1}\hspace{1.5em}Chen Huang\Mark{2}\hspace{1.5em}Chen Change Loy\Mark{3}}\\
	\Mark{1}CUHK-SenseTime Joint Lab, The Chinese University of Hong Kong\\
	\Mark{2}Robotics Institute, Carnegie Mellon University\\
	\Mark{3}School of Computer Science and Engineering, Nanyang Technological University\\	
	{\tt\small ly015@ie.cuhk.edu.hk~~chenh2@andrew.cmu.edu~~ccloy@ntu.edu.sg}
}


\maketitle

\appendix
\setcounter{table}{0}
\renewcommand{\thetable}{S\arabic{table}}
\setcounter{figure}{0}
\renewcommand{\thefigure}{S\arabic{figure}}

\section{Network Architecture}
\begin{figure*}[h]
	\centering
	\includegraphics[width=\linewidth]{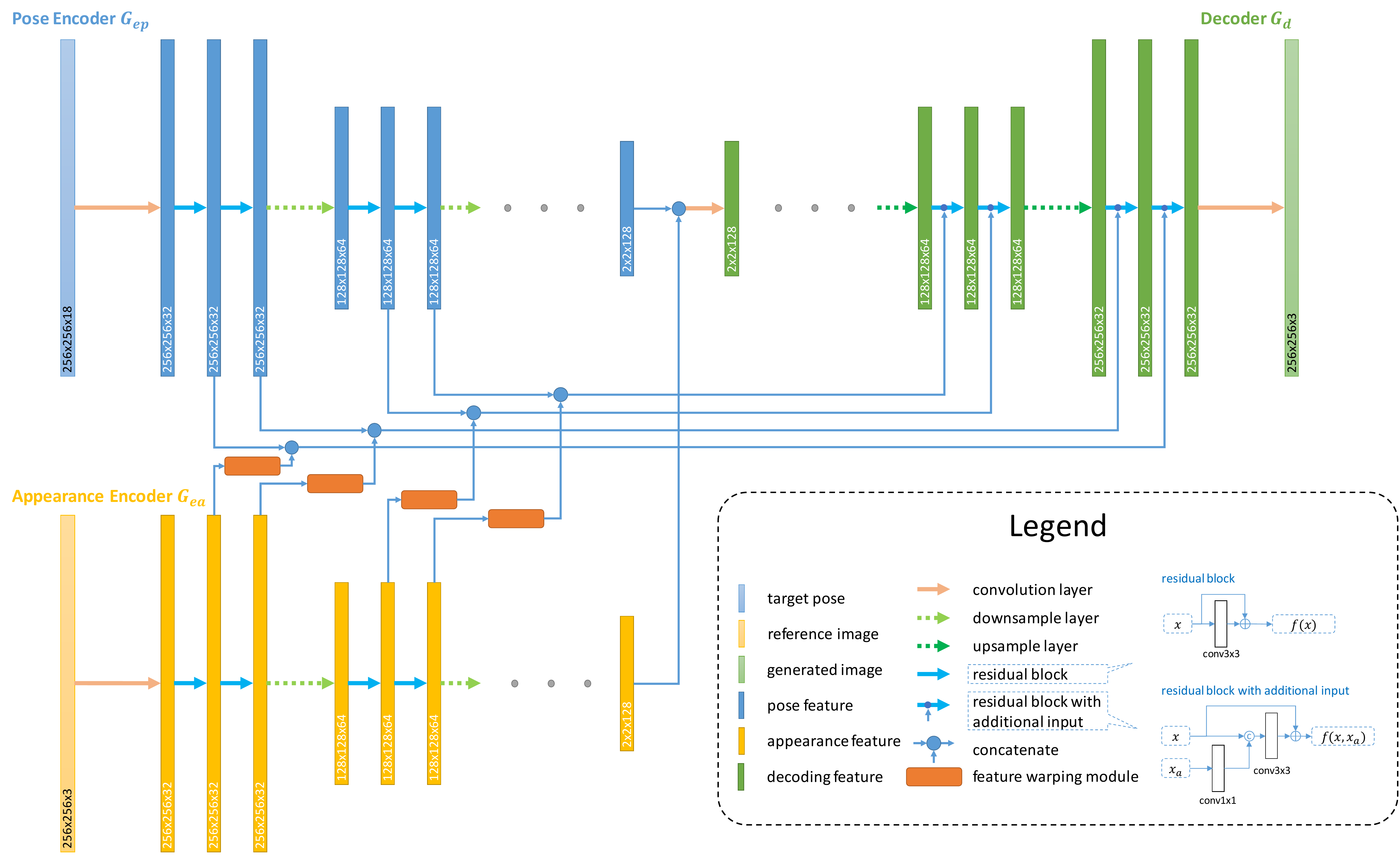}
	\caption{Detailed image generator network architecture.}
	\label{fig:supp_network}
\end{figure*}
Fig.~\ref{fig:supp_network} illustrates the detailed network architecture of our image generator described in Sec.4.1 in the main paper. There are $N=7$ feature levels in the network, where the number of channels at each level increases linearly from 32 to 128. We apply flow guided feature warping at the first 5 feature levels because higher level features with small spatial resolution are not location-sensitive. At the bottleneck, the encoded pose features and warped appearance features are directly concatenated.
\section{More Qualitative Results on DeepFashion Dataset}
\subsection{3D Appearance Flow Regression}
\begin{figure*}[t]
	\centering
	\includegraphics[width=\linewidth]{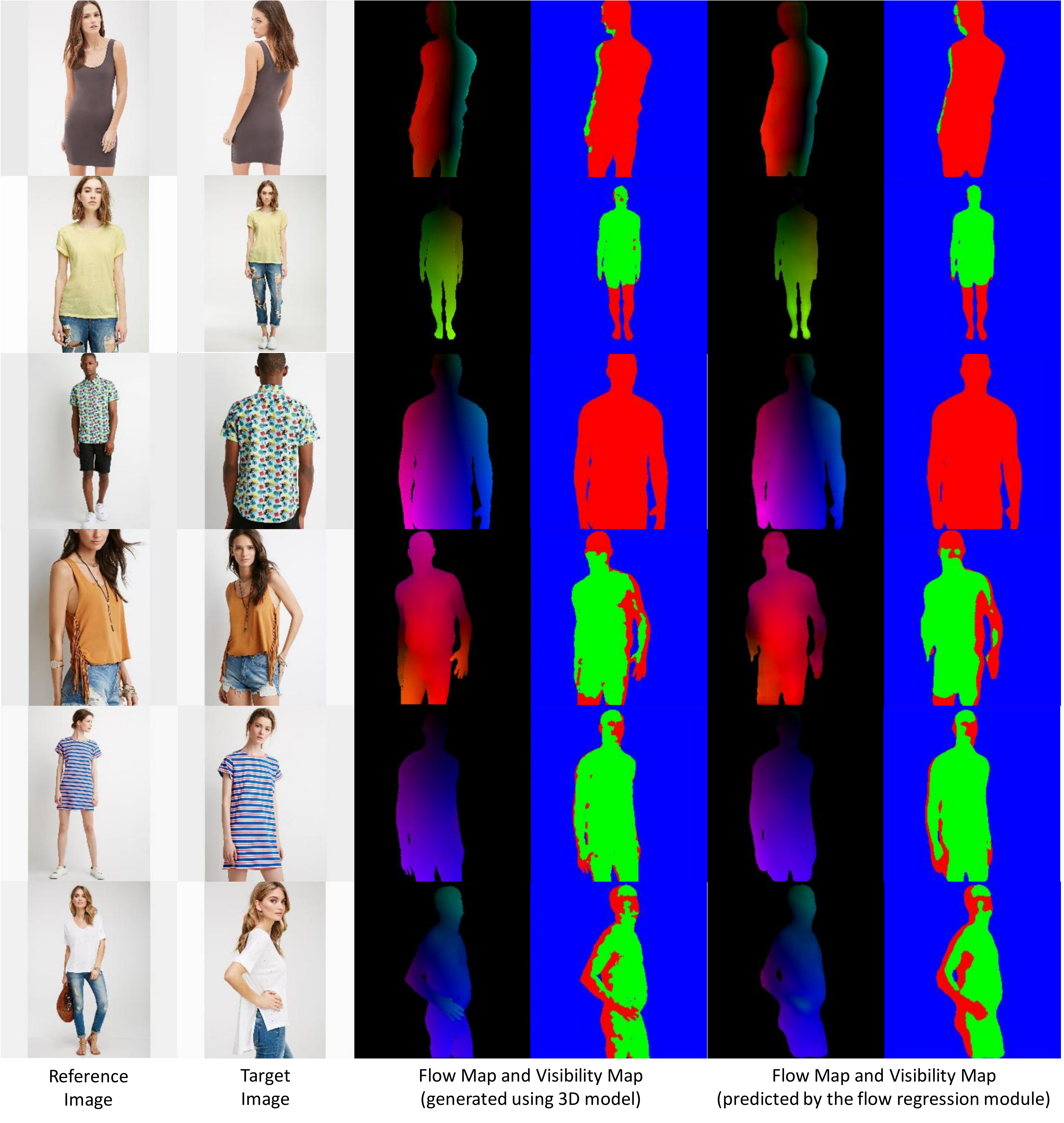}
	\caption{Examples of 3D appearance flow map and visibility map: generated ground-truth (middle) and prediction from our flow regression module (right).}
	\label{fig:supp_flow}
\end{figure*}
Fig.~\ref{fig:supp_flow} shows more examples of the 3D appearance flow map and visibility map. The predicted appearance flow maps are accurate (close to the ground-truth) regardless of diverse pose and viewpoint changes. While the predicted visibility map can accurately identify invisible regions caused by self-occlusion (\eg,~the 1st and 3rd rows) or out-of-field of view (\eg,~the 2nd row).

\subsection{Comparison with Previous Works}
\begin{figure*}[t]
	\centering
	\includegraphics[width=0.7\linewidth]{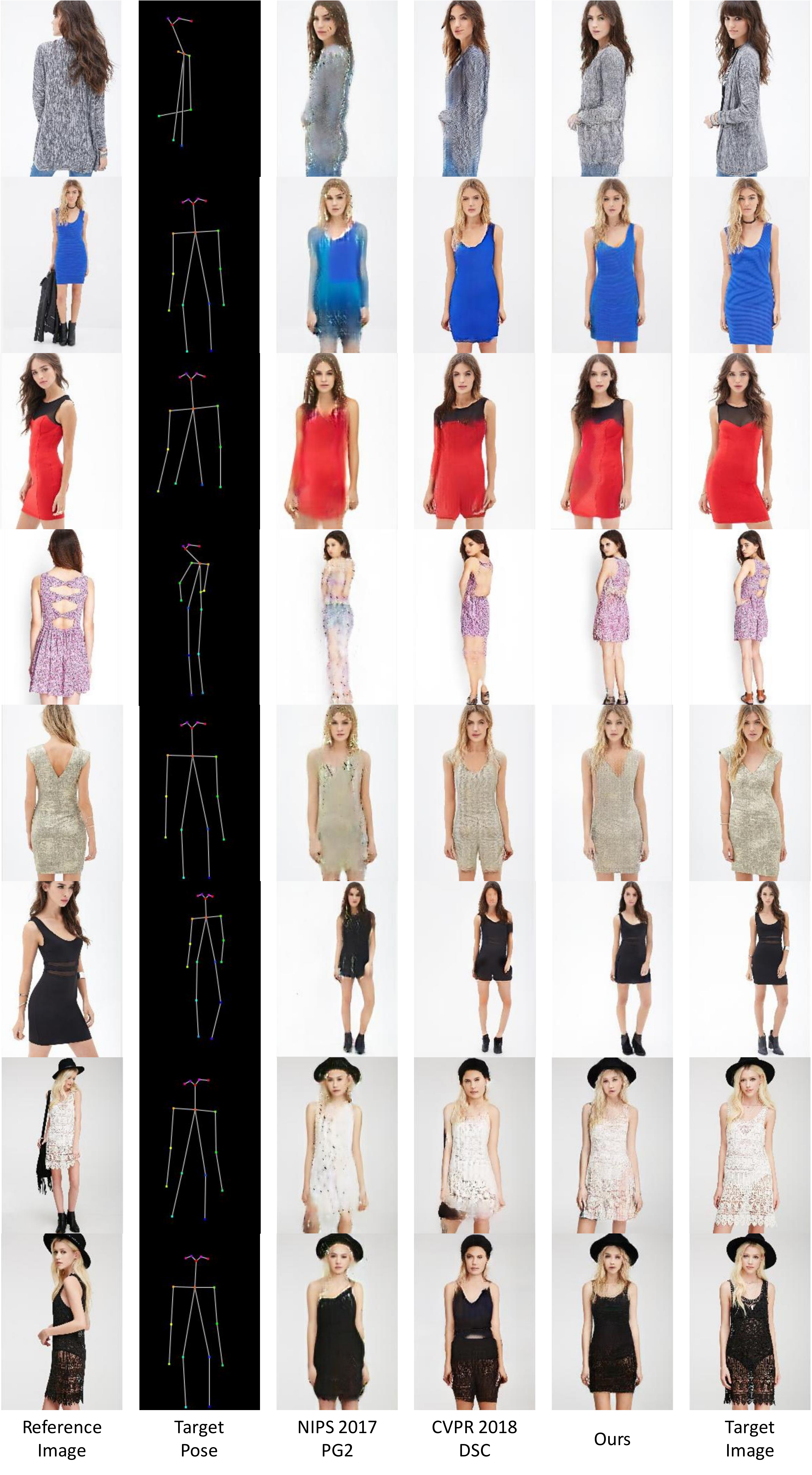}
	\caption{Qualitative comparison between our method and previous works (PG2~\cite{ma2017pose} and DSC~\cite{siarohin2018deformable}).}
	\label{fig:supp_compare}
\end{figure*}
Fig.~\ref{fig:supp_compare} shows more qualitative comparisons between our method and previous works~\cite{ma2017pose, siarohin2018deformable}. Results show that our method is able to generate more realistic images and better preserve the key appearance attributes.

\subsection{Arbitrary Pose Transfer}
\begin{figure*}[t]
	\centering
	\includegraphics[width=\linewidth]{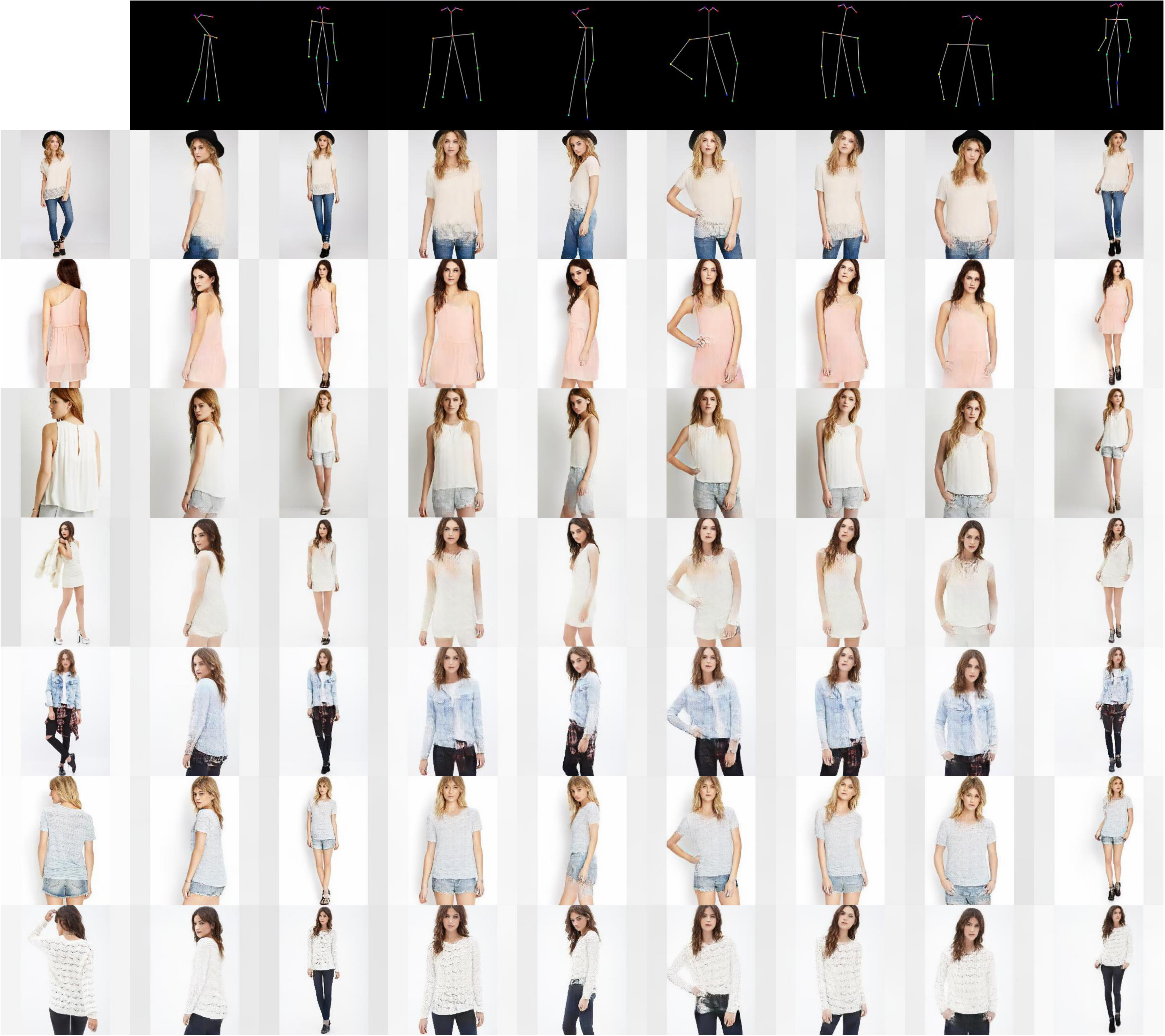}
	\caption{Arbitrary pose transfer results. Each image is synthesized using the leftmost reference image and the corresponding target pose.}
	\label{fig:supp_matrix}
\end{figure*}
We further test our method on transferring a reference image to arbitrary poses, and show the results in Fig.~\ref{fig:supp_matrix}. In each row, the leftmost image is the reference image, which is used to synthesize new images in different target poses. It is good to see that our method can effectively generalize to diverse and difficult human poses.

\subsection{Failure Case Analysis}
\begin{figure*}[t]
	\centering
	\includegraphics[width=0.8\linewidth]{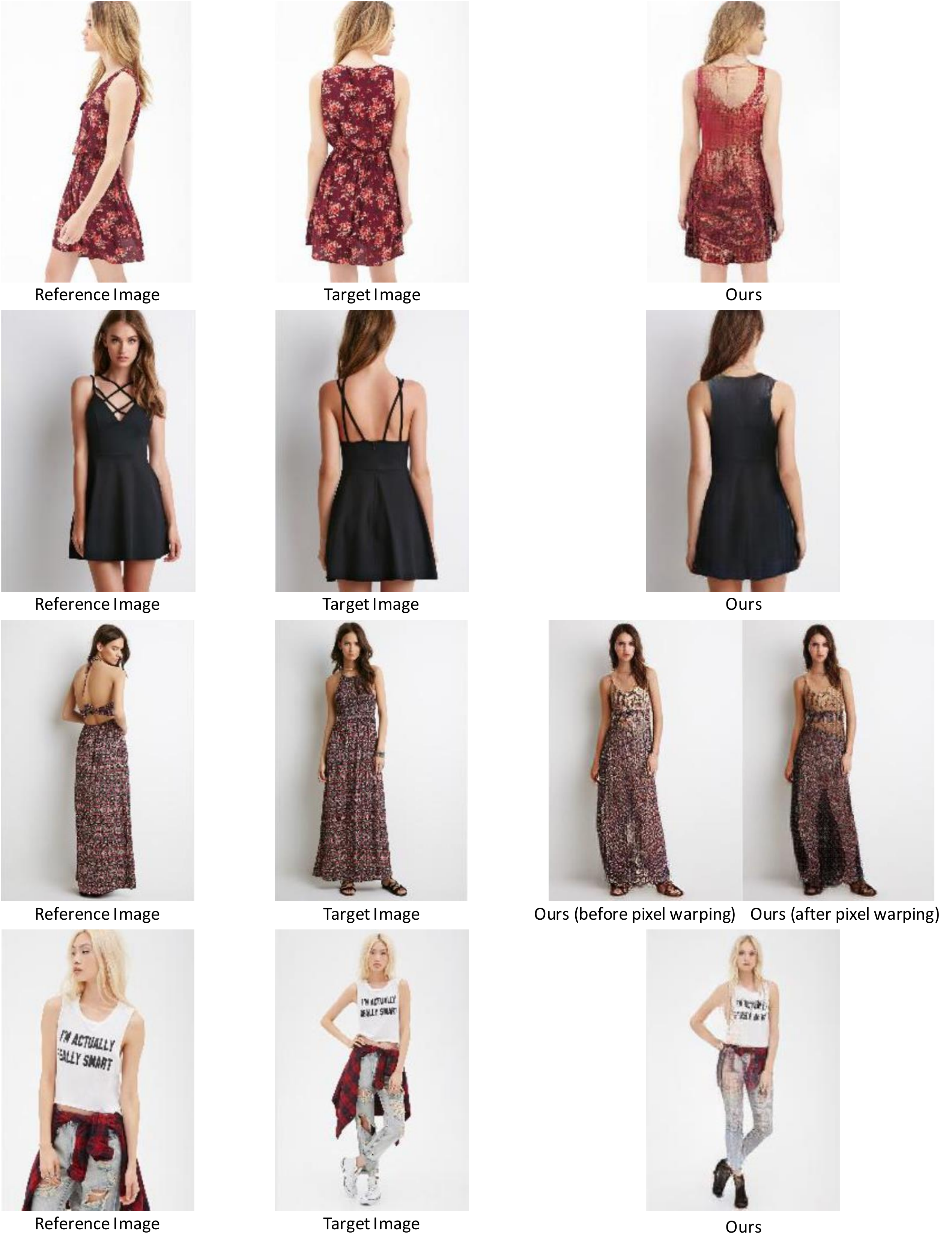}
	\caption{Example failure cases}
	\label{fig:supp_failure_case}
\end{figure*}
Fig.~\ref{fig:supp_failure_case} illustrates some failure cases of our method. The 1st and 4th rows show that our method has difficulty synthesizing some complex textures or special clothing layout,~\eg,~coat wrapped around the person's waist. Larger training data are expected to improve out model's ability to hallucinate rare textures. In the second row, our method fails to correctly infer the backside of a person from her frontal appearance. Although our generated result is also plausible, the shoulder part seems not so compatible with the frontal image. We think more training data can help enrich the expressiveness of our image generator. The third row shows one failure case when our pixel warping module tries to blend an inconsistent reference image region into the generated image, which is again due to the large front-back pose discrepancy.
\section{Experiments on Market-1501 dataset}
\begin{table}[h]
	\centering
	\caption{Quantitative results on Market-1501 dataset.}
	\label{table:supp_market1501}
	\small
	\begin{tabular}{@{ }c@{ }|c c@{ }c@{ } c}
		\hline
		Model & SSIM & Masked SSIM & IS & Masked IS\\
		\hline
		PG2~\cite{ma2017pose} & 0.253 & 0.792 & \textbf{3.460} & 3.435\\
		DSC~\cite{siarohin2018deformable} & 0.290 & 0.805 & 3.185 & 3.502\\
		\hline
		w/o. dual encoder& 0.290 & 0.868 & 2.918 & 3.568\\
		w/o. flow& 0.292 & 0.869 & 2.905 & 3.664\\
		w/o. visibility& 0.296 & 0.872 & 3.193 & \textbf{3.730}\\
		w/o. pixel warping& 0.303 & 0.873 & 2.986 & 3.699\\
		Ours full& \textbf{0.308} & \textbf{0.874} & 3.010 & 3.700\\
		\hline
	\end{tabular}
	\vskip -0.2cm
\end{table}
\begin{figure*}[t]
	\centering
	\includegraphics[width=\linewidth]{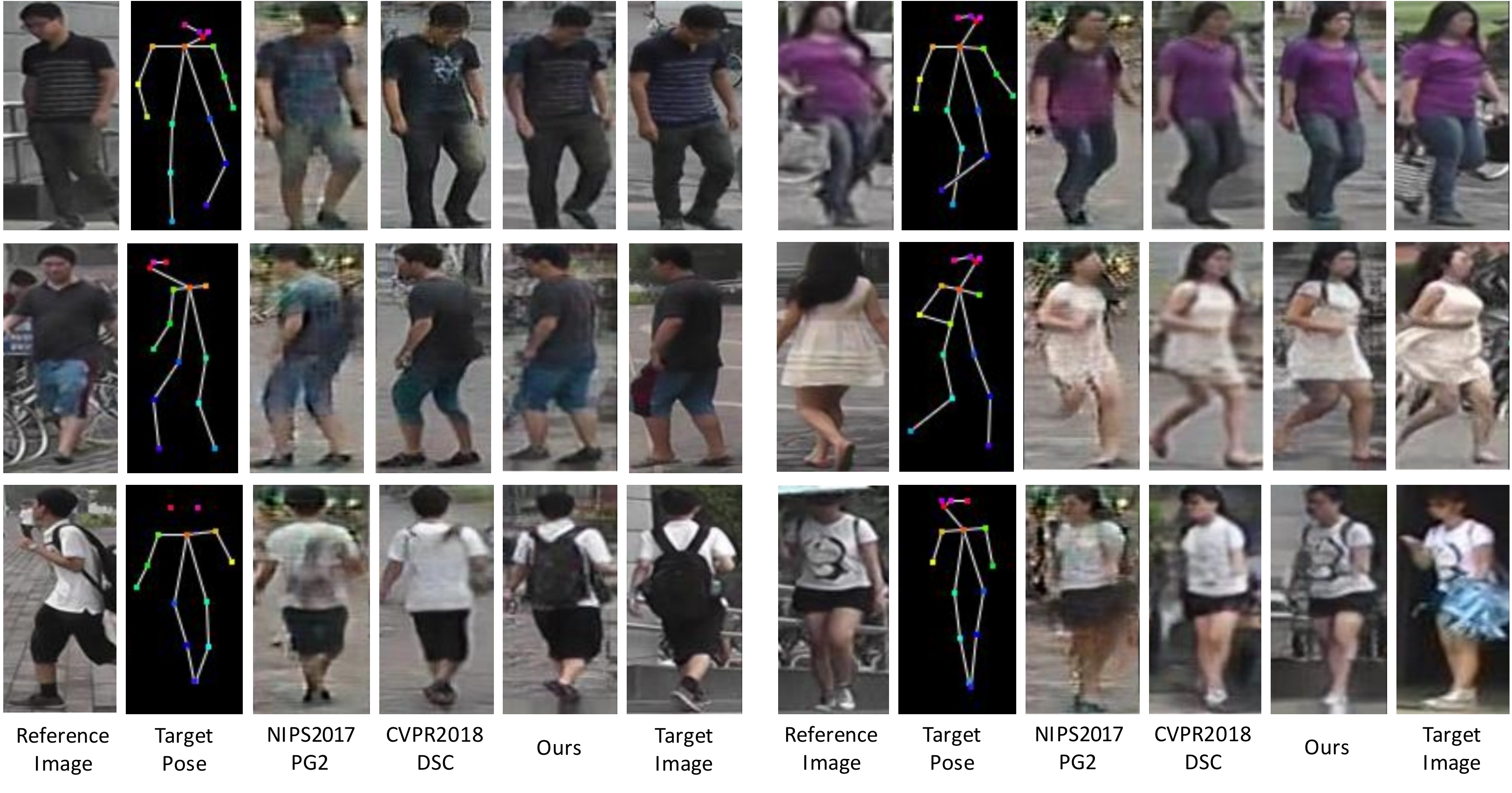}
	\caption{Qualitative results on Market-1501 dataset.}
	\label{fig:supp_market1501}
\end{figure*}

We further evaluate our model on the Market-1501 dataset~\cite{zheng2015scalable}, which consists of 32,668 surveillance images of 1,501 persons. Images in this dataset have a lower resolution of $128\times64$ pixels, but contain more diverse poses and complex backgrounds in comparison to images in DeepFashion dataset~\cite{liu2016deepfashion}. We follow the data splits in~\cite{siarohin2018deformable} and select 263,631 pairs for training and 12,800 pairs for testing. We modify the U-Net architecture of our image generator to reduce to $N=5$ levels due to the lower image resolution.

We show the quantitative results on Market-1501 in Table~\ref{table:supp_market1501}, and visualize some generated results in Fig.~\ref{fig:supp_market1501}. Our method achieves pretty strong results when compared to state-of-the-art baselines~\cite{ma2017pose, siarohin2018deformable}, and is able to generate higher quality details such as the backpack and clothing pattern.

{\small
\bibliographystyle{ieee_fullname}
\bibliography{supplement}
}